\documentclass[lettersize,journal]{IEEEtran}
\usepackage{amsmath,amsfonts}
\usepackage{algorithm}
\usepackage{array}
\usepackage[caption=false,font=normalsize,labelfont=sf,textfont=sf]{subfig}
\usepackage{color}
\usepackage{textcomp}
\usepackage{soul}
\usepackage{stfloats}
\usepackage{url}
\usepackage{verbatim}
\usepackage{graphicx}
\usepackage{cite}
\usepackage{threeparttable} % table note

\usepackage{multicol}
\usepackage{multirow}
\usepackage{booktabs}

\usepackage{algcompatible}
\algblockdefx{FORP}{ENDFP}[1]
  {\textbf{for }#1 \textbf{do in parallel}}
  {\textbf{end for}}
  
\usepackage{tabularx} 
\makeatletter
\newcommand{\multiline}[1]{%
  \begin{tabularx}{\dimexpr\linewidth-\ALG@thistlm}[t]{@{}X@{}}
    #1
  \end{tabularx}
}
\makeatother

\usepackage{colortbl} % table in blue

\usepackage{booktabs} % To thicken table lines

\begin{document}

\title{CoTV: Cooperative Control for Traffic Light Signals and Connected Autonomous Vehicles using Deep Reinforcement Learning}

\author
{          
 ~Jiaying~Guo,~\IEEEmembership{Student Member,~IEEE},
  ~Long~Cheng,~\IEEEmembership{Senior Member,~IEEE},
 and Shen Wang,~\IEEEmembership{Member,~IEEE}

\thanks{Jiaying Guo is with the School of Computer Science, University College Dublin, Ireland, E-mail: jiaying.guo@ucdconnect.ie }
\thanks{Long Cheng is with the School of Control and Computer Engineering, North China Electric Power University, Beijing 102206, China. E-mail: lcheng@ncepu.edu.cn}
\thanks{Shen Wang is with the School of Computer Science, University College Dublin, Ireland, E-mail: shen.wang@ucd.ie }
}

\markboth{Journal of \LaTeX\ Class Files,~Vol.~xx, No.~x, August~xxxx}%
{Shell \MakeLowercase{\textit{et al.}}: A Sample Article Using IEEEtran.cls for IEEE Journals}


\maketitle

\begin{abstract}
The target of reducing travel time only is insufficient to support the development of future smart transportation systems. To align with the United Nations Sustainable Development Goals (UN-SDG), a further reduction of fuel and emissions, improvements of traffic safety, and the ease of infrastructure deployment and maintenance should also be considered. Different from existing work focusing on the optimization of the control in either traffic light signal (to improve the intersection throughput), or vehicle speed (to stabilize the traffic), this paper presents a multi-agent Deep Reinforcement Learning (DRL) system called CoTV, which \underline{Co}operatively controls both \underline{T}raffic light signals and Connected Autonomous \underline{V}ehicles (CAV). Therefore, our CoTV can well balance the achievement of the reduction of travel time, fuel, and emissions. In the meantime, CoTV can also be easy to deploy by cooperating with only one CAV that is the nearest to the traffic light controller on each incoming road. This enables more efficient coordination between traffic light controllers and CAV, thus leading to the convergence of training CoTV under the large-scale multi-agent scenario that is traditionally difficult to converge. We give the detailed system design of CoTV and demonstrate its effectiveness in a simulation study using SUMO under various grid maps and realistic urban scenarios with mixed-autonomy traffic.
\end{abstract}


\section{Introduction}

% \IEEEPARstart{T}{he} core contribution of this paper is to demonstrate using DRL to train two different types of agent (TL CAV) to achieve better traffic.

% Notice that we have to highlight the extra contributions we've made compared to our VNC poster (mainly on \textit{deployment/robustness}):
% \begin{itemize}
%     \item mixed-autonomy:
%     \item scalability (traffic efficiency \& DRL training):
%     \item communications (delay \& loss):
% \end{itemize}

\IEEEPARstart{D}eveloping the next generation Intelligent Transportation Systems (ITS) is one of the key ways to achieve the United Nations Sustainable Development Goals (UN-SDG)\cite{desa2016transforming}. In particular, firstly, sustainable traffic requires higher efficiency to reduce enormous monetary losses caused by excessive traffic delays. Secondly, more eco-friendly driving should be encouraged to decrease fuel consumption and gas emissions (mainly CO\textsubscript{2}). Thirdly, traffic safety is one of the key indicators for sustainable traffic, inherently, which should be enhanced by avoiding potential collisions to save lives. Last but not least, to achieve those sustainable traffic goals, easy-to-deploy ITS infrastructure is critical.

\begin{figure}[ht!]
    \centering
    \includegraphics[width=0.9\linewidth]{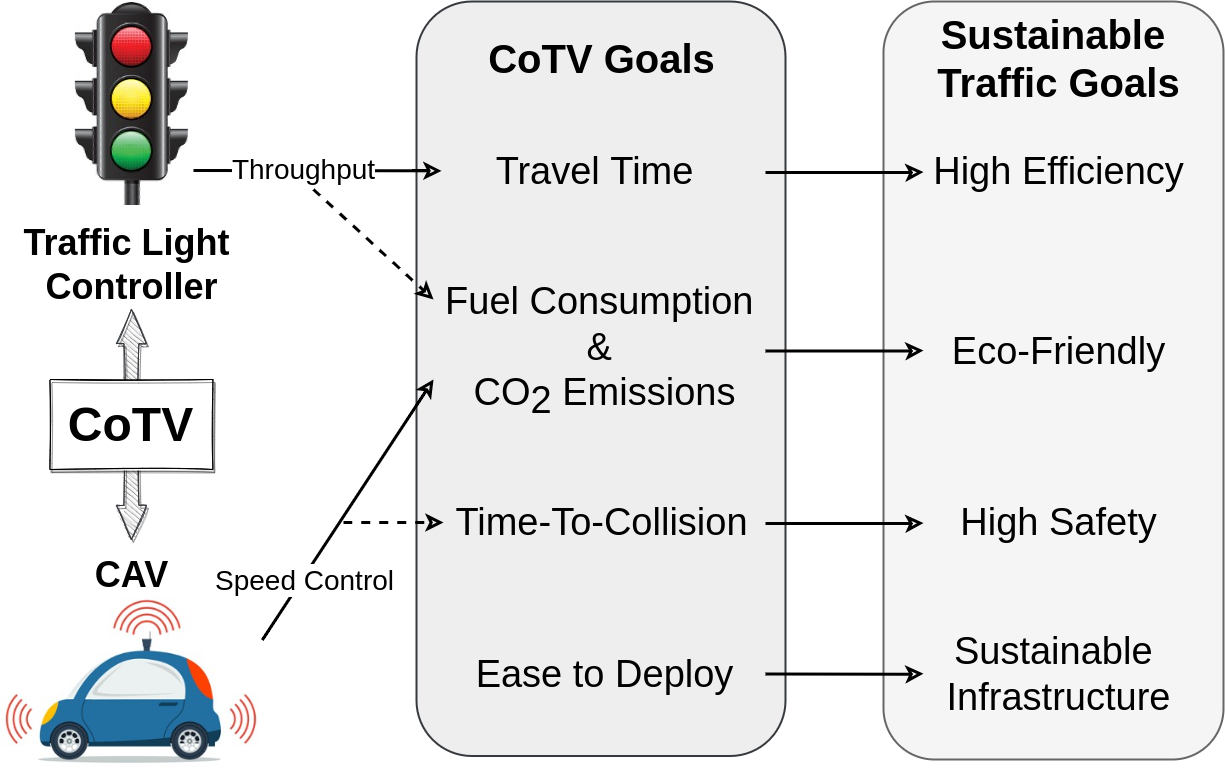}
    \caption{The illustration of the motivation and goals of our proposed system CoTV. Traditionally, traffic light controllers can increase the intersection throughput, thus reducing the travel time and fuel. While CAV adjusts its speed to reduce the fuel, thus maintaining a safe time gap to its surrounding traffic. Our CoTV coordinates these two different types of agents to achieve a more comprehensive set of the goals of sustainable traffic.}
    \label{fig:moti}
\end{figure}

Most existing research in sustainable urban traffic control adjusts either traffic light signals or vehicle speed. Traffic light signal controllers dynamically select the best timing plan according to the real-time traffic. As shown in Fig.\ref{fig:moti}, this can directly increase the intersection throughput, thus reducing travel time as well as energy consumption and emissions. CAV can proactively control vehicles' acceleration, as shown in Fig. \ref{fig:moti}, to achieve more stable traffic nearby with relatively higher driving velocity (i.e., lower fuel consumption and gas emissions) and keep a safe distance \cite{shetty2021safety} from the surrounding traffic (i.e., longer time-to-collision). Recent research from the transportation domain attempts to explore the potential of joint control for both traffic light signals and vehicle speed. Methodologies used in such research include mixed-integer linear programming \cite{yu2018integrated}, the enumeration method and the pseudo-spectral method \cite{xu2018cooperative}. However, these methods may not perform well in realistic traffic scenarios because their deterministic traffic control decisions are insufficient to deal with a fast-changing urban environment\cite{di2021survey}.

Unlike the aforementioned traditional methods, many researchers nowadays have demonstrated the great potential of DRL in solving traffic control challenges under complex urban scenarios. For instance, inspired by the traditional traffic signal control method MaxPressure \cite{varaiya2013max}, PressLight \cite{wei2019presslight} can achieve even better traffic efficiency improvements under various urban scenarios using DRL. Moreover, the DRL-based traffic signal control can also reduce the waiting time of specific vehicles in emergency situations in which traffic condition varies quickly\cite{benedetti2021application}. On the other hand, efficient and effective CAV speed control can stabilize traffic in many complex and changing road scenarios using DRL\cite{wu2017flowframework}, which is traditionally infeasible using optimization-based controllers. However, there is a lack of research using DRL for the joint control of both urban intersection signals and vehicle speed. This DRL-based joint control is challenging due to the difficulty of designing a proper cooperation scheme for two different agent types (i.e., traffic light controllers and CAV). Moreover, the unpredictability of urban mixed-autonomy traffic makes it even harder to converge within a reasonable number of training iterations.

To overcome the limitations mentioned above, we propose CoTV: a multi-agent DRL-based system that can cooperatively control traffic light signals and CAV. CoTV well balances the advantages of both traffic light controllers and CAV to achieve more sustainable traffic, as shown in Fig.\ref{fig:moti}. Concretely, the contributions of our work are as follows:

\begin{itemize}
    \item \textbf{Effective cooperation schemes between CAVs and traffic light controllers.} Different from the methodology in the literature on Multi-Agent Reinforcement Learning (MARL) for traffic control, instead of using action-dependent design \cite{wei2019survey} (i.e., the action of one agent depends on the action of other agents in the shared environment), our cooperation schemes rely on the exchange of states between agents within the range of one intersection, including the traffic light controller and approaching CAVs. This so-called ``action-independent MARL" \cite{chu2019multi} can work in our CoTV as the objective of traffic light controller and CAV for the traffic improvement are inherently complementary (i.e., rather than overlapping: all improving travel time or reducing fuel). Thus, CoTV takes advantage of the simplicity of ``action-independent MARL" design on DRL training and keeps the effectiveness of CoTV in improving traffic under various scenarios. The cooperation schemes of CoTV are shown to facilitate training convergence, which is challenging for independent MARL that does not include any cooperation (either state or action). Specifically, our CoTV using Proximal Policy Optimization (PPO) \cite{schulman2017proximal} obtains up to 30\% reduction in both travel time and fuel consumption \& CO\textsubscript{2} emissions under varying CAV penetration rates.
    \item \textbf{Scalable to complex urban scenarios by avoiding cooperation with excessive CAV agents.} Compared with controlling all possible CAVs using MARL, the traffic light controller in our CoTV selects the closest CAV to the intersection on each incoming road as the CAV agent. This idea is inspired by platooning can increase intersection throughput \cite{lioris2016doubling}, as the leading vehicle in a certain road has the great potential to form a platoon with the rest vehicles on the same road. We also demonstrate that compared with coordinating all CAVs (CoTV*), CoTV does not compromise efficiency improvement while significantly reducing the training time and resources used.
    \item \textbf{Efficient communication exchange schemes between CAV and traffic light controllers.} The amount of state information exchanged between CAV and traffic light controllers is low enough. As shown in Fig.\ref{fig:system}, the communication schemes are designed to exchange the speed, acceleration, and location of CAVs and the current signal phase of traffic light controllers to each other. The information exchange requires less than 100 Kbps transmission rate, which can be achieved using Vehicle-To-Vehicle (V2V) and Vehicle-To-Infrastructure (V2I) communication infrastructure. The wireless communication technology IEEE 802.11p supports a bandwidth of 3 Mbps to 20 Mbps \cite{singh2019tutorial}.
    % small -> (1. what are the information exchanged < 100Kbps; 2. V2V V2I IEEE802.11p bandwidth 3-20Mbps)
\end{itemize}

This paper extends our previous work \cite{guo2021poster} to control traffic light signals and CAV cooperatively using DRL. The improvements include: 1) The system framework of CoTV is designed for addressing scalability issues, resulting in the significantly reduced number of CAV agents controlled. 2) The state and reward for agents are simplified by removing redundant traffic information. Therefore, the amount of information exchanged among agents is reduced to ease the deployment of CoTV. 3) The testing scenarios are extended from a small grid map to more realistic urban scenarios. 4) We demonstrate the robustness of CoTV under different CAV penetration rates. 5) As an important requirement of achieving sustainable traffic, the effectiveness of CoTV in enhancing traffic safety is evaluated by time-to-collision \cite{lee1976theory}. 6) Two other common MARL approaches, action-dependent and independent, are compared with the action-independent MARL of our CoTV in terms of policy training and traffic improvements.

\section{Related Works}

% This section introduces the existing work in optimising the control of traffic light signal and vehicle speed, respectively, as well as the cooperative control of them both. The recent studies of mixed-autonomy traffic are also included.
This section overviews the recent related work and highlights the gaps that our CoTV attempts to fill. In particular, it firstly focuses on the research in either traffic light signal control or vehicle speed control. Secondly, it discusses the recent research in the joint control of both agents. Lastly, it summarizes the practicability of deploying existing work in mixed-autonomy and its impact on traffic efficiency and safety.

% Some important papers are: \cite{farah2018infrastructure}

\subsection{Control for Either Traffic Light Signals or Vehicle Speed} 

Most existing research in sustainable urban traffic control adjusts either traffic light signals or vehicle speed.
Sydney Coordinated Adaptive Traffic System (SCATS) \cite{sims1980sydney} is one of the earliest and most widely applied traffic light signal control systems. It can dynamically select the best signal plan from a list of pre-defined candidates that can potentially achieve better intersection throughput by improving green time efficiency. 
Varaiya \cite{varaiya2013max} proposed a traffic light signal control scheme named MaxPressure, which was proven to maximize the throughput of the entire road network, with each traffic light controller receiving local traffic information.
On the other hand, the field experiments in \cite{stern2018dissipation} prove that the speed control of CAV can stabilize traffic and is beneficial to reduce braking times and fuel consumption.
Green Light Optimal Speed Advisory (GLOSA) system guides CAV to adjust its speed according to the current traffic signal phase and the remaining distance to its approaching intersection \cite{suzuki2018new}. Therefore, the more smooth acceleration/deceleration of CAVs can further reduce fuel consumption and CO\textsubscript{2} emissions.
However, these traffic control optimization approaches rely on deterministic formulations to make dynamic traffic problems tractable. These deterministic formulations remain static in ever-changing traffic and thus may not be flexible enough to improve realistic traffic.
% These traffic control optimization approaches utilize limited input features to make traffic problems tractable and efficient in designated scenarios. However, decisions of these rule-based models are deterministic based on assumptions. Many factors in real-world traffic (i.e., varying traffic demand and heterogeneous road conditions) cannot be considered without being described in these traffic models. Traditional rule-based methods can not generalize well in ever-changing real-world traffic.

DRL has been used to cope with complex traffic environments, promising better urban traffic. 
PressLight \cite{wei2019presslight} is a DRL-based model using Deep Q-learning (DQN). It collects local real-time traffic information inspired by the traditional method MaxPressure \cite{varaiya2013max} while achieving more improvement on traffic efficiency than MaxPressure.
Wu et al. \cite{wu2017flowframework} extended the field experiments in \cite{stern2018dissipation} using the Trust Region Policy Optimization (TRPO) method for training CAVs in a simulated experiment. The used DRL-based vehicle speed controller surpasses traditional optimization controllers on traffic improvement.
Various scenarios using CAV had been tested in \cite{vinitsky2018benchmarks}, including road merging and unsignalized intersections. The DRL-based speed control of CAV can optimize the vehicle trajectory of the whole trip and reduce the risk of collision all the time.
Compared to traditional optimization methods with deterministic solutions, DRL methods, which are used in our proposed CoTV, learn from trial-and-error in the interaction with the environment to train different optimal policies under various traffic scenarios, which is more capable of performing adaptive traffic control and generalizing well under fast-changing urban road scenarios.

\subsection{Joint Control for Traffic Light Signals and Vehicle Speed}

Traditional optimization-based methods have been attempted to jointly control traffic light signals and vehicle speed.
Yu et al. \cite{yu2018integrated} developed mixed-integer linear programming for optimizing vehicle trajectories and traffic signals simultaneously at isolated intersections. 
The phase sequence and duration of traffic light signals are coordinated with vehicle arriving time to the intersections.
A two-level model for traffic light controllers and CAV was proposed using the enumeration method and the pseudo-spectral method \cite{xu2018cooperative}. The first level is applied to coordinate CAV and traffic light controllers to minimize travel time, and the second level is used to regulate CAV trajectory to reduce fuel consumption.
The same system targets were adopted in the cooperative optimization model \cite{tajalli2021traffic}. The model uses a mixed-integer non-linear program, which originally has high computational complexity.
% Yu et al. \cite{yu2018integrated} developed mixed-integer linear programming for optimizing vehicle trajectories and traffic signals simultaneously in a unified framework. The optimization model coordinates signal phase sequence and duration with vehicle arrival time at the intersections to determine the exact vehicle trajectory during several signal plan cycles, which is impractical to apply for more realistic urban traffic. A two-level optimization model for traffic light controllers and CAV \cite{xu2018cooperative, tajalli2021traffic} separates the cooperative control problems in different reduced updating intervals. Optimal traffic signal timing is first calculated using the enumeration method, while speed control of CAV is more frequently updated by the pseudo-spectral method under the calculated traffic signal \cite{xu2018cooperative}. However, these models have been only investigated at an isolated intersection. Eco-Approach and Departure systems with adaptive traffic signal control \cite{hao2018eco} are more practical in real-world traffic.

To the best of our knowledge, DRL methods for the joint control of traffic light signals and CAV have not been well studied.
The joint control using DRL suffers many challenges, commonly in multi-agent systems \cite{hernandez2019survey}: (1) Every agent, traffic light controller, or CAV, proactively interacts with the same environment simultaneously, causing a non-stationary environment to bring more uncertainty on training convergence. (2) A large number of agents cause scalability issues due to an exponential increase in the computational cost of joint action. (3) The reward of agents can assess the system at a different area scale in the environment: individual, regional, or global. Reward design is critical for DRL agents due to the high correlation to achieving system goals. For example, traffic light controllers explicitly coordinate traffic around intersections, and each CAV mainly affects its surrounding traffic. 
The proposed model in this paper attempts to overcome these difficulties and utilize the advantage of DRL methods to control traffic light signals and CAV cooperatively.

\subsection{Efficiency and Safety for Mixed-Autonomy Traffic}

The development of CAV is thriving in both academia and industry, which is expected to improve traffic. However, the deployment must experience a gradual mixed transition from introductory, established, to prevalent \cite{olstam2020approach} with the growth of CAV penetration rate. Existing work presents that CAV mixing in traffic still brings uncertainty. Mixed-autonomy experiments on motorways were conducted in \cite{postigo2021effects}, simplified from intersections with conflicting traffic movements. Similar work was tested in single-lane facilities, where CAV can enhance traffic safety by keeping a larger gap from the surrounding vehicles \cite{sharma2021assessing}. However, a low penetration rate (less than 10\%) causes more conflicts in urban scenarios \cite{gueriau2020quantifying}. On the other hand, CAV has the potential to improve traffic efficiency but cannot guarantee a higher average speed than traditional vehicles depending on the network type and traffic conditions. \cite{wei2019mixed} conducted experiments in a ring scenario, showing that the CAV penetration rate greater than 20\% allows all vehicles to reach higher speeds and stabilize the flow. The penetration rate in 20\% to 40\% is possible to result in the near-maximum improvements \cite{gueriau2020quantifying}. Overall, a high penetration rate of CAV can bring traffic efficiency and safety improvement on mixed-autonomy traffic in various scenarios.

This work advances the state-of-the-art in assessing DRL-based mix-autonomy control under dynamic urban road scenarios with multiple intersections. Moreover, our system CoTV chooses only a small fraction of CAVs that cooperate with traffic light controllers, which have great potential to guide the rest of the vehicles. This makes the deployment of CoTV practical and easy-to-scale.
% However, using the same threshold of TTC, there are no road conflicts in various scenarios with the CAV penetration rates from 0\% to 100\% \cite{niroumand2020joint}.
%Quang Tran and Bae \cite{quang2020proximal} implemented experiments at a non-signalized intersection. The larger penetration rate of CAVs brings more positive effects on driving efficiency, improves average speed, and reduces delay. However, fuel consumption fluctuates when the penetration rate is less than 70\%. Close to full CAVs can slightly decrease fuel consumption. 
%At the isolated signalized intersections, the experiment results in \cite{tajalli2021traffic} illustrate the increase in CAV penetration rate can reduce travel time and queue length. 
% However, the improvement in traffic performance has fluctuated with different penetration rates \cite{du2021coupled}.  And higher average speed brings more fuel consumption and CO2 emissions. 

\section{System Overview}

% In this section, we first outline our system goals. Then the system components, two types of agents, traffic light controllers and CAV, are presented with the design of its action, state, and reward, respectively. Moreover, we demonstrate the system procedure and highlight the efficient communication schemes as well as the method to avoid excessive CAV agents for cooperative control. The DRL algorithm used is also be discussed.

This section explains the design of our system CoTV. Firstly, we outline the system design goals. Then, the system components (i.e., traffic light controllers and CAV) are presented with the design of their action, state, and reward. The cooperation schemes between the two types of agents using Vehicle-To-Everything (V2X) communications are elaborated as well. Thirdly, we explain the training process of CoTV using PPO, during which parameter sharing is applied for all agents in the same type to perform the learned policy. Additionally, we also present the consideration of the ease of deployment in designing CoTV. The code of this study is open-sourced \footnote{See https://github.com/Guojyjy/CoTV}.

% system goals -> system components -> system work flow -> RL design -> deployment considerations

\subsection{System Design Goals}

The proposed model CoTV aims to achieve the following goals, which are also shown in Fig.\ref{fig:moti}:

\begin{itemize}
    \item \textbf{Reduced travel time}: 
    Travel time is the metric that end road users care about the most. Our system should reduce the travel time of a vehicle with a given route. This goal is traditionally achieved by traffic light signal control that can increase intersection throughput.
    \item \textbf{Lower fuel consumption and CO\textsubscript{2} emissions}:
    Sustainable traffic goals encourage eco-friendly driving behaviors. This goal is traditionally achieved by the speed control of CAV that can stabilize traffic flow. Smart traffic light control can also partly contribute to achieving this goal by reducing the number of stop-and-go.
    \item \textbf{Longer time-to-collision}:
    Safety is a crucial consideration in sustainable traffic system design. Reducing the risk of collision can be achieved by maintaining a longer time-to-collision \cite{lee1976theory}, with sufficient time to moderately decelerate. CAV can proactively keep a safe distance from the surrounding traffic. Thus, ITS using CAV has the potential to achieve higher traffic safety.
    \item \textbf{Easy to deploy}:
    Our system CoTV requires a V2X communication infrastructure to support communication exchange over the cooperative control. Meanwhile, scalability issues should be addressed with the increasing number of agents. Efficient communication schemes among traffic light controllers and reduced CAV agents are the key to achieving this goal.
\end{itemize}

\subsection{System Components}

\begin{figure}[ht]
    \centering
    \includegraphics[width=\linewidth]{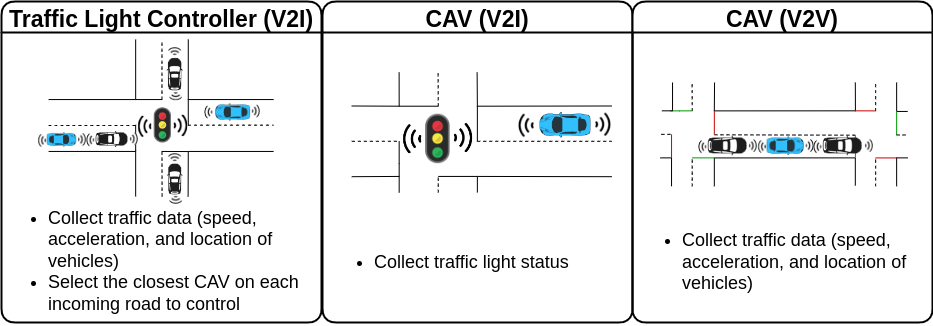}
    \caption{V2X communication schemes in CoTV showing how traffic light controllers and CAVs use V2I and V2V. This implements state exchange and cooperative control. CAV agents of CoTV are highlighted in blue.}
    \label{fig:v2x}
\end{figure}

Our proposed system assumes that all vehicles are connected, including CAV and Human-Driven Vehicles (HDV) (details can be found in Table \ref{tab:vehicle}). The V2X communication is also assumed perfect without no packet loss and no latency. The main components of CoTV: traffic light controllers and CAV, as shown in Fig.\ref{fig:system}. The design of action, state, and reward for them are described as follows, while the V2X communication schemes involved are shown in Fig.\ref{fig:v2x}:

\begin{figure*}[htb]
    \centering
    \includegraphics[width=0.9\linewidth]{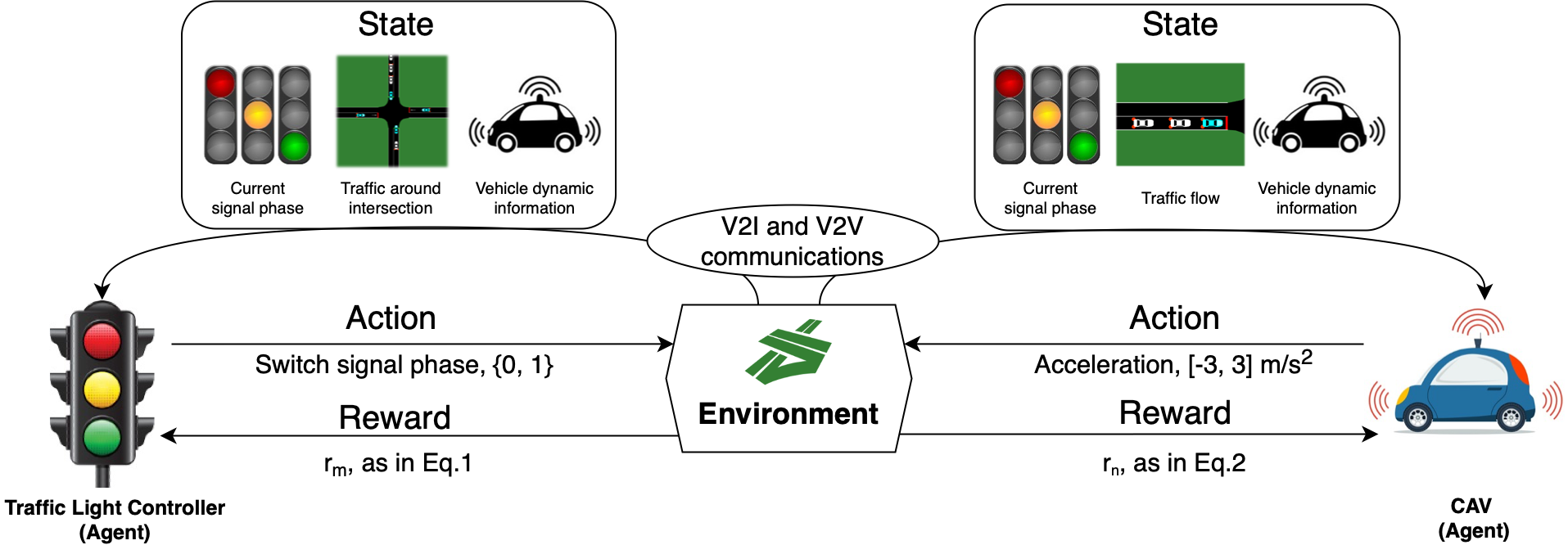}
    \caption{The DRL design of CoTV. Two types of agents, traffic light controllers and CAV interact with the environment according to the state information exchanged via V2X communications.}
    \label{fig:system}
\end{figure*}

\subsubsection{Traffic light controller}
\begin{figure}[ht]
  \includegraphics[width=\linewidth]{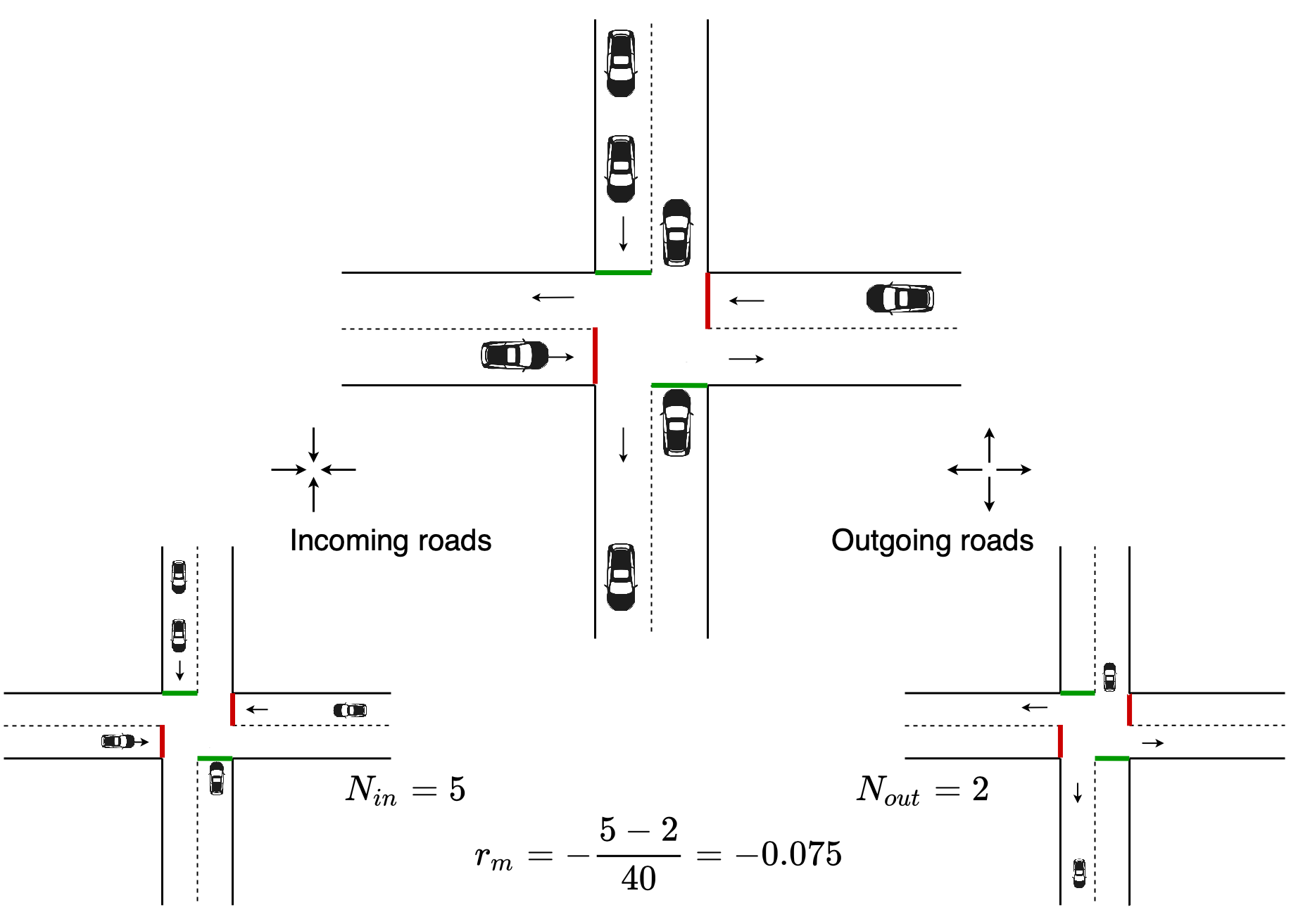}
  \caption{The illustration of the reward of a traffic light controller $r_m$, assuming the maximum road capacity $c = 40$.}
  \label{fig:tl_reward} 
\end{figure}

\begin{itemize}
    \item \textbf{Action}: We limit the action of traffic light controller to a binary set, where ``1" represents switching to the next phase for the next timestep while ``0" means to keep the current phase unchanged. As opposed to other common action definitions in the literature, such as phase selection\cite{chu2019multi}, the phase switch \cite{wei2018intellilight} we choose is more manageable for the model training process.
    \item \textbf{State}: The state of traffic light controller involves three parts: the current signal phase, traffic on the roads that this traffic light controller coordinates, and the status of the closest vehicle to the intersection on each incoming road. As shown in Fig.\ref{fig:system}, of all three parts of the traffic light controller's state, the information of the last two parts is acquired by using the V2I communications infrastructure illustrated in Fig.\ref{fig:v2x}. The road traffic is presented by the number of vehicles on each road coordinated by the traffic light controller. These roads are divided into incoming roads and outgoing roads. The last part of the state includes speed, acceleration, distance to the intersection, and the road name where it is located for the closest vehicle to the intersection on each incoming road.
    \item \textbf{Reward}: The reward is the penalty of intersection pressure inspired by \cite{wei2019presslight, varaiya2013max}. Intersection pressure is defined as the difference between the sum of the number of vehicles on the incoming roads $N_{in}$ and the sum of the number of vehicles on the outgoing roads $N_{out}$. Then the intersection pressure is normalized by the maximum road capacity $c$ to improve DRL training. The maximum road capacity $c$ indicates the maximum number of vehicles on a single road in the given road network. It is calculated by dividing the length of the longest road by the minimum space required for a vehicle (i.e., the length of a single vehicle plus the minimum distance between two adjacent vehicles).
    % as shown in Eq.(\ref{equ:c}). 
    % \begin{equation}
    %     c = \frac{D_{max}}{L_{v}+gap}
    %     \label{equ:c}
    % \end{equation}}
    The reward of a certain traffic light controller $r_m$, Eq.(\ref{equ: R-TL}) becomes
    \begin{equation}
        r_m = - \frac{N_{in} - N_{out}}{c} 
        \label{equ: R-TL}
    \end{equation}
    We also illustrate the above reward in Fig.\ref{fig:tl_reward}. The reward function is formulated to reduce travel time, one of the system goals, by increasing intersection throughput. Minimizing intersection pressure encourages vehicles to pass through the intersection quickly while considering the remaining capacity in the outgoing roads, thus improving green light efficiency and throughput \cite{varaiya2013max}. We also simplify the calculation of intersection pressure without considering traffic movements (correspondence between incoming and outgoing roads) compared with \cite{varaiya2013max,wei2019presslight}. Therefore, CoTV can be easily applied in various urban scenarios with multi-directional roads. Besides, we avoid using other common reward definitions in the current literature, such as queue length and waiting time \cite{wei2019survey, wei2018intellilight}, which is precarious in different traffic flow conditions even without the influence of traffic lights.  
\end{itemize}

\subsubsection{CAV}
\begin{figure}[ht]
    \centering
    \includegraphics[width=\linewidth]{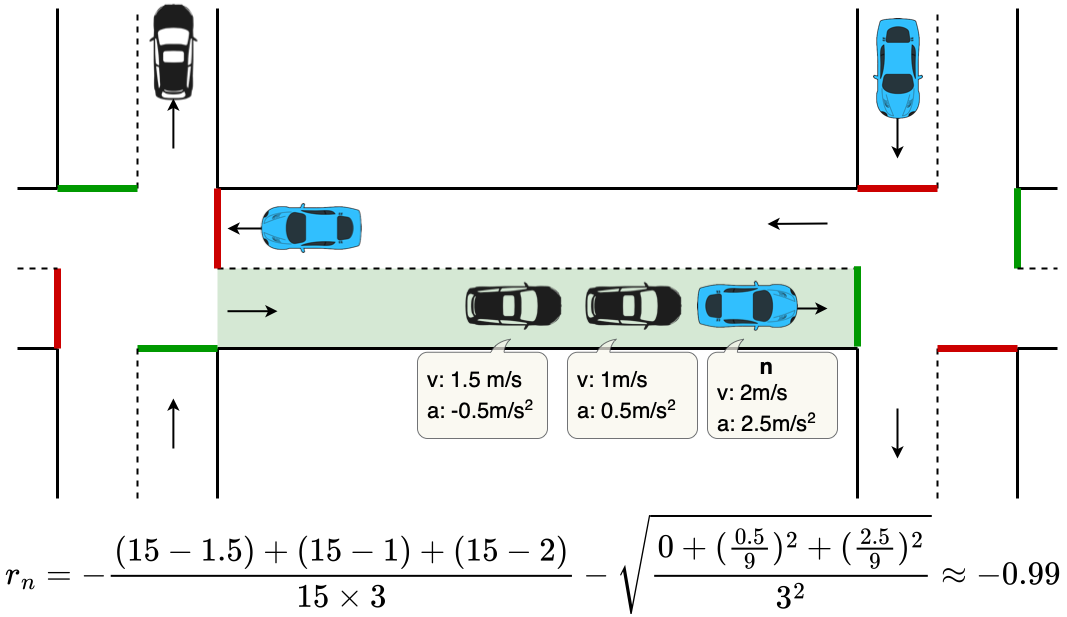}
    \caption{Illustration of the CAV reward $r_n$, assuming the maximum speed limit $v^*=15$, and the vehicle's maximum acceleration $a^*=9$. CAV agents of CoTV are highlighted in blue.}
    \label{fig:cav_reward}
\end{figure}

\begin{itemize}
    \item \textbf{Action}: The action is set to be consistent with the literature \cite{wu2017flow}, which is a continuous action space to represent the CAV acceleration in the range of $[-3m/s^2, 3m/s^2]$.
    \item \textbf{State}: The state explicitly includes speed and acceleration for itself and the vehicle preceding the CAV immediately, the distances to the preceding vehicle and the approaching intersection, and the current signal status of the approaching traffic light controller. The CAV agent can receive information from the vehicles on the same road and the approaching traffic light controller using V2V and V2I communication, as shown in Fig.\ref{fig:v2x}.
    \item \textbf{Reward}: The reward is penalized by the deviation of average speed $v$ from the maximum speed limit $v^*$, plus the Euclidean norm of acceleration $a$ after the normalization using the vehicle's maximum acceleration $a^*$, as shown in Fig.\ref{fig:cav_reward}. Speeds and accelerations in the reward are that of all vehicles $K$ located on the same road as the CAV agent. The reward of certain CAV agent $r_n$, Eq.(\ref{equ: R-AV}) becomes
    \begin{equation}
    \begin{aligned}
        & r_n = r_1 + r_2, \\
        & r_1 = - \frac{\sum_{j \in K} (v^* - v_j)}{v^* \times |K|},\,  v_j \leq v^*, \\
        & r_2 = - \sqrt{\frac{\sum_{j \in K} (\frac{a_j}{a^*})^{2}}{|K|^{2}}},\,  
        a_j = \begin{cases}
        0,  a_j < 0 \\
        a_j,  a_j \geq 0
        \end{cases}
        \label{equ: R-AV}
    \end{aligned}
    \end{equation}
    The first term of the reward function $r_1$ encourages a higher average vehicle velocity but keeps it within the maximum speed limit. In this speed range, higher speed increases fuel economy, and potential collisions due to excessive speed can be avoided. Moreover, collisions can generally be avoided as they often lead to a significant decrease in the speed of many following vehicles blocked by such collisions (i.e., such training episodes will be discarded due to low reward value). The second term of the reward function $r_2$ stabilizes acceleration to reduce fuel consumption, while also inducing a large time gap between adjacent vehicles \cite{sharma2021assessing} for enabling high-speed collision-free driving. Our reward function of CAV agents encourages better speed control, thus facilitating cooperative control of CoTV to achieve the reduction of fuel consumption and CO\textsubscript{2} emissions and the improvement of traffic safety.
\end{itemize}
%  \frac{1}{|K|} 

\subsection{Training Process}

% Algorithm 1 presents the training process and the outcome is to get the policy functions for traffic light agents $M$ and CAV agents $N$, $\pi_{TL}$ and $\pi_{CAV}$. The trained policy $\pi$ is expected to guide agents to select the appropriate action $a$ in a certain state $s$ for maximizing the accumulated value of reward $r$. The collected trajectory is sampled to update the traffic light and CAV policy through gradient descent in each iteration.

Algorithm 1 presents the training process of CoTV, and the outcome is the policy functions for traffic light agents $M$ and CAV agents $N$, $\pi_{TL}$ and $\pi_{CAV}$. The trained policy $\pi$ is expected to guide agents to select one appropriate action $a$ from the action set in a certain state $s$ for maximizing the accumulated value of reward $r$.  We predefine the termination condition as the number of training iterations $I$. In each iteration, there are $E$ episodes running in parallel, and each episode lasts $H$ timesteps. DRL trajectory data $\tau$ is collected per simulation timestep in each episode to extend the training batch $B$, and then sampled $K$ times to update the traffic light and CAV policy parameter $\theta_{TL}$ and $\theta_{CAV}$ through gradient descent. Specifically, the traffic light controllers of CoTV select the closest CAV to the intersection on each incoming road as the CAV agent, as in Line 10 of Algorithm 1. These CAV agents have the potential to increase intersection throughput by forming a platoon with the rest vehicles on the same road \cite{lioris2016doubling}. The communication exchange schemes for the cooperative control of CoTV occur in the agent receiving state, in Line 13 of Algorithm 1. Traffic light controllers and CAV exchange information with each other, involving the current signal status of traffic light and speed, acceleration, and location of CAV.

We choose PPO algorithm \cite{schulman2017proximal} for the following reasons. The PPO algorithm has the advantage of being easy to implement and achieving monotonic reward improvement. DQN is the common algorithm to train traffic light controllers \cite{wei2018intellilight, wei2019presslight}, which is efficient in the discrete actions (e.g., a binary set of signal phase adjustment). Whereas DQN does not perform well on continuous actions (e.g., vehicle acceleration of any real number within a certain range) \cite{lillicrap2015continuous}. In contrast, PPO can be applied for scenarios with discrete actions or continuous actions. On the other hand, compared with traffic light signals that have a pre-defined phase sequence, the initial driving behavior of DRL-controlled CAV has lots of unreasonable stop-and-go and standstill. The constrained policy update of PPO aims to improve reward monotonically, which is more stable to train CAV and better than Asynchronous Advantage Actor-Critic used in \cite{chu2019multi}. Although TRPO can also constraint the policy update, PPO is easy to implement and simpler to sample data, which helps the cooperation of traffic light controllers and CAV.

When interacting with the environment, CoTV applies parameter sharing \cite{wu2017flowframework} to all agents of the same type in the multi-agent DRL system, which can converge the training process faster and benefit from shared experience, especially in large-scale applications \cite{chen2020toward}.

% \hl{This paper could be a good example on how to present RL-based algorithm/mathematical notations: https://ieeexplore.ieee.org/abstract/document/9641753}
\begin{figure}[htbp]
  \label{alg:workflow}
  \renewcommand{\algorithmicrequire}{\textbf{Require:}}
  \renewcommand{\algorithmicensure}{\textbf{Ensure:}}
  \begin{algorithm}[H]
    \caption{Training Process of CoTV using PPO}
    \begin{algorithmic}[1]
      \REQUIRE 
      \STATE Obtain the set of traffic light agents to control, $M$
      \STATE Set the number of episodes in parallel to $E$, and  the time horizon for each episode to $H$
      \STATE Initialize the policy parameter for one type of agent, $\theta_{TL}$ for traffic light controllers and $\theta_{CAV}$ for CAV, through parameter sharing
    %   \STATE Initialize the set of all actions for traffic light controllers and CAV, $A_{TL}$ and $A_{CAV}$
      \STATE Initialize sample batch $B = \emptyset$ % to store agent trajectory for policy update
      \STATE Set the number of epochs for mini-batch updates in one iteration as $K$
      \ENSURE
      \FOR{iteration = 1,2,...,$I$}
        \FORP{episode = 1,2,...,$E$} % parallel
            % \STATE Reset the environment
            \FOR{timestep $h$ = 0,1,...,$H$}
                \FOR{each traffic light agent $m$ in $M$}
                    % \STATE Receives the state $s_{h_m}$ 
                    \State \multiline{%
                    Add the closest CAV $n$ to the intersection on each incoming road to the CAV agent set $N$}
                \ENDFOR
                % \FOR{each CAV agent $n$ in $N$} % parallel
                    % \STATE Receives the state $s_{h_n}$
                % \ENDFOR
                \FOR{each agent $i$ in $M + N$}
                    % \State \multiline{%
                    % Selects the action $a_{h_i} \sim \pi(a|s_{h_i}, \theta), a \in A_{TL}, A_{CAV}$}
                    % \State \multiline{%
                    % Receives the reward $r_{h_i}(s_{h_i},a_{h_i})$}
                    % \State \multiline{%
                    % Stores the trajectory $(s_{h-1_i}, a_{h_i}, s_{h_i}, r_{h_i})$ in the sample batch $B$}
                    \STATE Run policy $\pi_{TL}$ or $\pi_{CAV}$ in the environment
                    % \State \multiline{%
                    \STATE Collect trajectories $\tau = (s_{h-1_i}, a_{h_i}, s_{h_i}, r_{h_i})$
                    \STATE Extend $B$ with $\tau$
                \ENDFOR
            \ENDFOR
            \STATE  Compute advantage estimates $\hat{A}_1,... \hat{A}_H$
        \ENDFP
        \State \multiline{%
        Update $\theta_{TL}$ and $\theta_{CAV}$ in the policy $\pi_{TL}$, $\pi_{CAV}$ using advantage estimates $\hat{A}$, with $K$ epochs to sample mini-batches from $B$, and then reset $B = \emptyset$}
    \ENDFOR
    \end{algorithmic}
  \end{algorithm}
\end{figure}

\subsection{Considerations for ``easy-to-deploy"}
% \begin{itemize}
%     \item nearest: compared with all
%     \item agent coordination: via state, rather than action-dependent
%     \item amount of information exchanged: a full calculation process
%     \item vehicular communication: within the range of a single intersection.
% \end{itemize}
Firstly, CoTV is designed to be deployed in the major junctions of urban scenarios, which requires minimum upgrades to the existing adaptive traffic light systems (e.g., SCATS, SCOOTS, etc.). This deployment strategy covers broader arterial roads that carry the majority of traffic by the minimum possible number of intersection controllers. Lane-changing operations are not considered in the action space of CoTV agent design for simplicity. However, lane-changing operations are permitted in the evaluation of CoTV shown in Section V. Secondly, compared with controlling all possible CAVs with DRL, the traffic light controller of CoTV selects only the closest CAV to the intersection on each incoming road to cooperate, which can significantly reduce training time and resources used in the process thus alleviating scalability issues.
Meanwhile, the cooperation schemes among agents (i.e., the traffic light controller and the approaching CAV agents) only rely on the information exchange of states, not actions. This means the action for a certain agent is selected independently from other agents' actions. Therefore, CoTV avoids the exponentially increased complexity of joint actions for MARL using action-dependent design \cite{hernandez2019survey}.
Besides, the amount of information exchanged in CoTV is low enough compared with high-dimensional transmission data (i.e., image representations to describe traffic features) \cite{wei2018intellilight, liang2019deep}. Specifically, as shown in Fig.\ref{fig:v2x}, the information of CAV involves speed, acceleration, and location. Their size is estimated to be approximately 40 Bytes if encoded using float numbers. While traffic light controllers send their current signal phase, which is about 8 Bytes if using integer numbers for encoding. This information plus headers will still be less than 100Kbps. This transmission demand is met by the V2I and V2V communications infrastructure \cite{singh2019tutorial} using IEEE 802.11p which is between 3 and 20 Mbps. Additionally, all the information exchanged using the vehicular network occurs within the range of a single intersection (i.e., the single-hop range that is about 300 meters), which can improve the robustness of CoTV instead of heavily relying on a large scale (i.e., using multi-hop transmission) of network conditions \cite{chu2019multi}.

\section{Evaluation Methodology}

In this section, we introduce the evaluation methodology, which includes the simulation settings, the metrics used for evaluation, and the overview of compared methods against our proposed CoTV.

\subsection{Simulation Scenarios}

The simulation platform used in this work is Simulation of Urban MObility (SUMO)\footnote{https://www.eclipse.org/sumo/}, which is one of the most widely used open-source microscopic traffic simulators. Our model design and implementation are based on FLOW\footnote{https://flow-project.github.io}, which provides DRL-related API to work with SUMO dynamically.

% For every scenario, the training process is terminated after 150 iterations. 18 episodes run in parallel in one iteration to simulate the dynamic traffic environment. Agent trajectories are collected during 720 timesteps in one episode. We set 1 timestep equal to 1 second.

We clarify some concepts relating to the time horizons. We set 1 \textit{simulation timestep} equal to 1 simulation second. One \textit{episode} refers to a full run of a single simulation scenario, which is set to 720 simulation timesteps. At the end of each \textit{iteration}, CoTV starts to update the parameters of the PPO algorithm used, after 18 episodes run in parallel. In total, we terminate the \textit{training process} of CoTV after 150 iterations.

For testing scenarios, firstly, we demonstrate the effectiveness of CoTV under a simple 1x1 grid map with a single intersection. Then, we show CoTV can be scalable to more consecutive intersections under a 1x6 grid map. Lastly, we validate the effectiveness of CoTV using a subset of the realistic urban scenario of Dublin city, Ireland. Table \ref{tab:traffic} summarizes the settings of traffic in each scenario.

\begin{table}[htbp]
  \begin{center}
    \caption{Traffic settings in the three test scenarios.}
    \label{tab:traffic}
    \begin{threeparttable}
        \begin{tabular}{p{1.5cm}|p{2.3cm}|p{2.3cm}}
          \toprule
          \textbf{Scenario} & \textbf{Traffic generation duration (seconds)} & \textbf{Total number of vehicles} \\
          \midrule
          1x1 grid\tnote{1} & 300  & 70 \\
          1x6 grid & 300 & 240 \\
          Dublin & 400 & 275 \\
          \bottomrule
        \end{tabular}
        \begin{tablenotes}
            \footnotesize
            \item[1] $a$x$b$ grid, $a$ is the number of row, $b$ is the number of column.
        \end{tablenotes}
    \end{threeparttable}
  \end{center}
\end{table}

\subsubsection{$1\times1$ grid map}
\begin{figure}[ht!]
    \centering
    \includegraphics[width=0.7\linewidth]{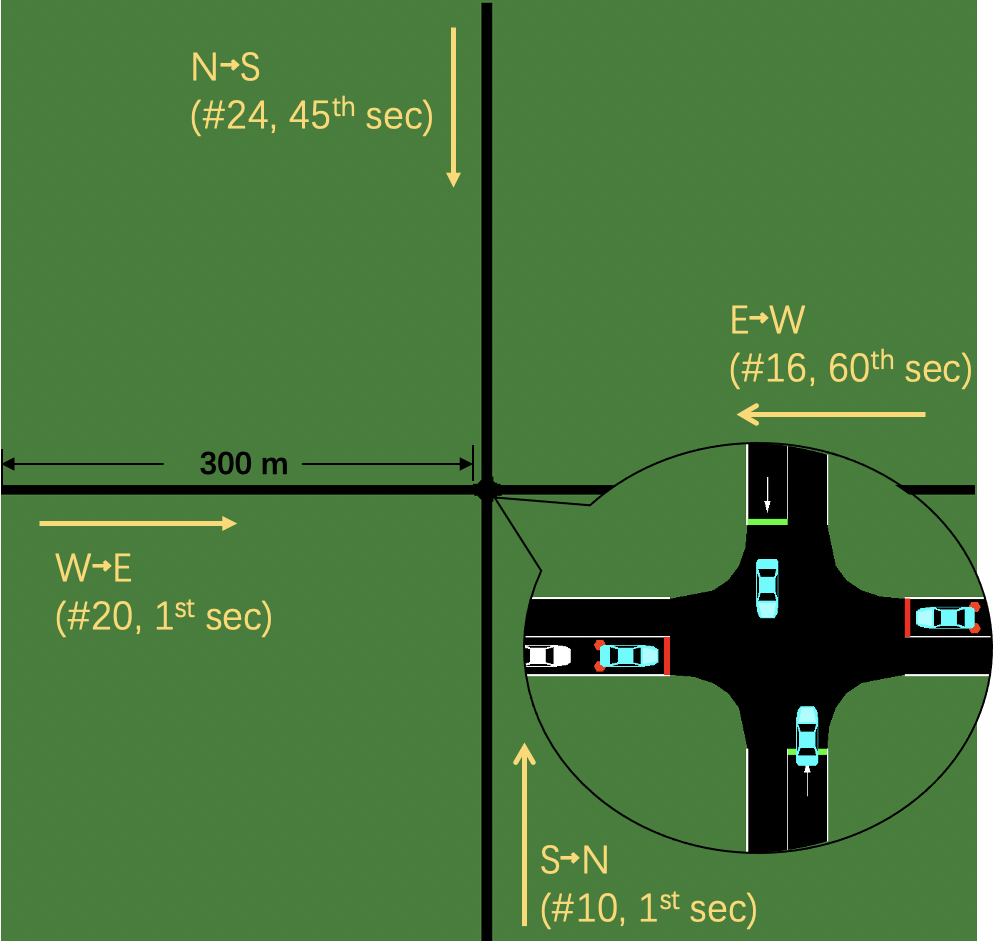}
    \caption{The settings of traffic generation for 1x1 grid scenario. For example, W$\rightarrow$E (\#20, 1st sec) means there are 20 vehicles sequentially generated from the first simulation second.}
    \label{fig:1tl}
\end{figure}

In our 1x1 grid map, each edge has two roads in opposite directions. To make this map closer to the real urban scenario, we set the road length as 300 meters and the maximum speed limit as 15 m/s (=54km/h). As shown in Fig.\ref{fig:1tl}, we generate different go-straight traffic flows in four directions: N→S (from north to south), S→N, W→E (from west to east), and E→W. This traffic generation method is inspired from \cite{chu2019multi}. The origin and destination of each vehicle are at the end of the road at the perimeter of the network. The vehicle generation duration for each flow is approximately 300 seconds. The traffic flows N→S and W→E are relatively heavier than the other two. Specifically, the traffic flow rates in the number of vehicles per hour per road are: 288 (N→S), 240 (W→E), 192 (E→W), and 120 (S→N), respectively. The two traffic flows, S→N and W→E, are generated at the beginning of each episode. Then, the N→S flow vehicles start to enter the network sequentially on the 45th second. After one minute, the traffic flow of E→W appears. The speed of each vehicle when entering the network is random. Thus, the total number of vehicles is 70 in the scenario with one intersection.
% Veh/h/ln is the common parameter to demonstrate the traffic demand, representing the number of vehicles generating per hour on each road.

\subsubsection{$1\times6$ grid map}
\begin{figure}[ht!]
    \centering
    \includegraphics[width=\linewidth]{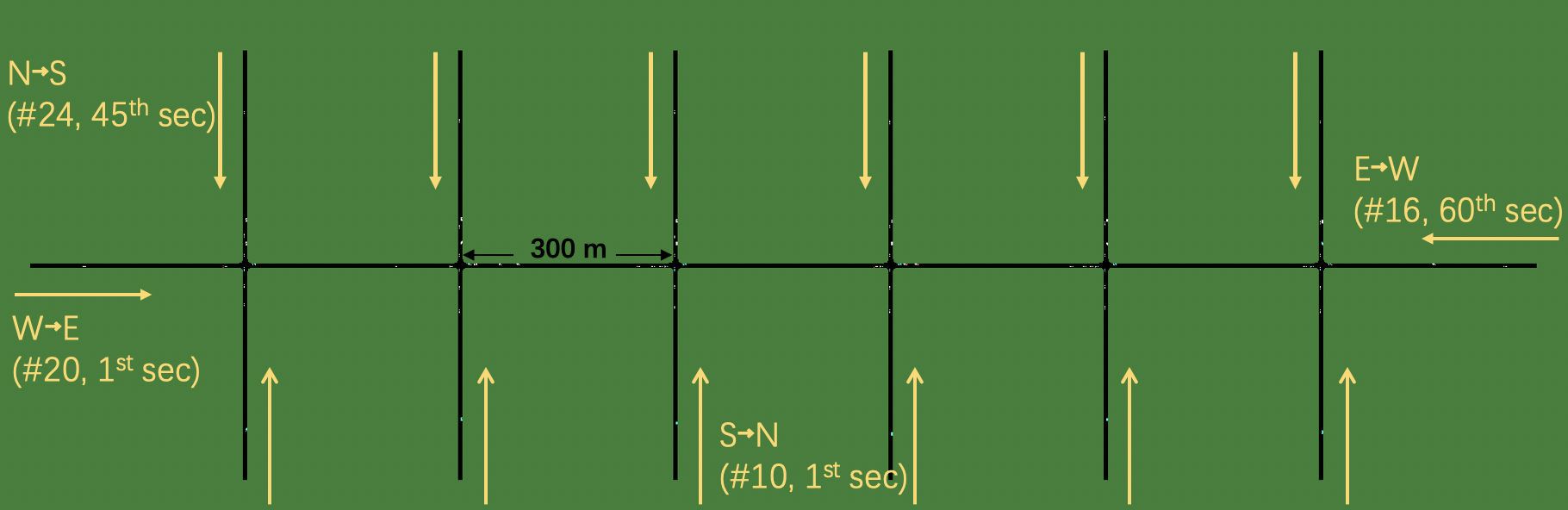}
    \caption{The settings of traffic generation for 1x6 grid scenario. The same settings (the number of vehicles generated, the simulation time to start traffic generation) apply for the traffic flow in the same direction.}
    \label{fig:6tl}
\end{figure}

The 1x6 grid scenario is shown in Fig.\ref{fig:6tl}, which contains six intersections extending the 1x1 grid map with 5 more consecutive intersections. The road setting and traffic flow configurations are similar to the settings of the 1x1 grid scenario. The increased vertical (N→S and S→N) roads are allocated the corresponding traffic flow. A total of 240 vehicles are generated in this scenario.

\subsubsection{Dublin map}

\begin{figure}[ht!]
    \centering
    \includegraphics[width=\linewidth]{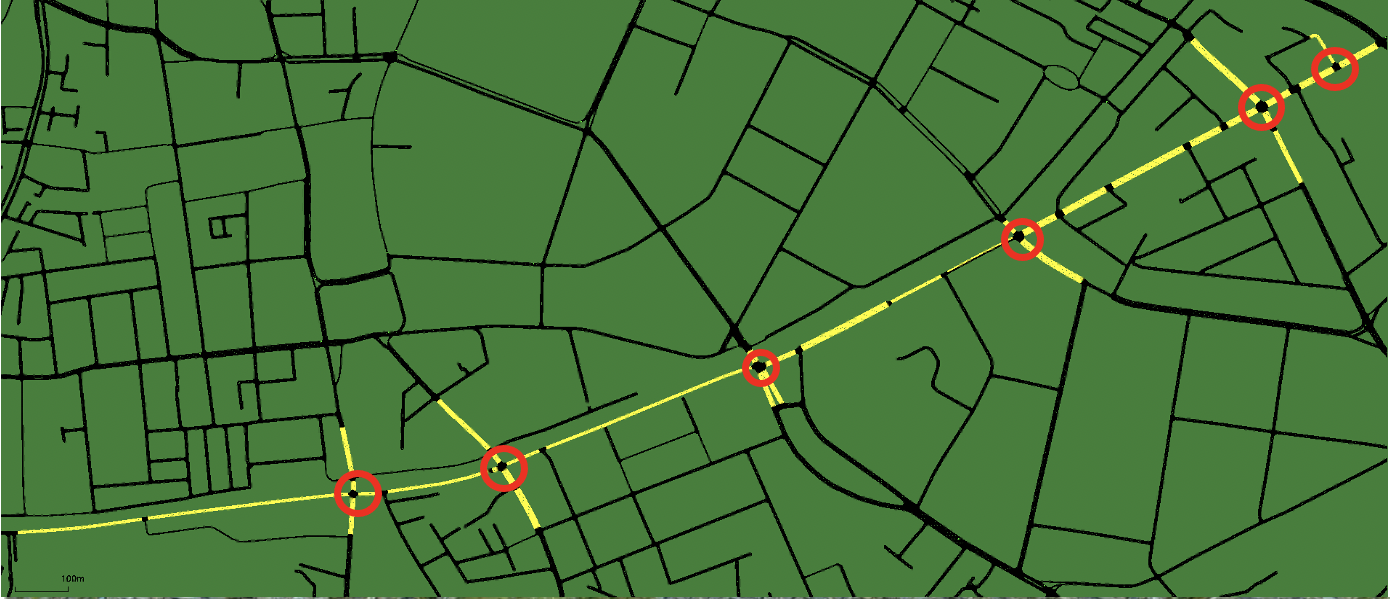}
    \caption{The selected six signalized intersections area in the city of Dublin (a regional road, R111, in South Dublin). The highlighted roads are our selected testing scenario (six intersections are highlighted using red circles).}
    \label{fig:Dublin}
\end{figure}

Fig.\ref{fig:Dublin} illustrates the selected six signalized intersections area in the city of Dublin. These intersections are the main ones connected by arterial roads, maximizing traffic improvement while considering ``easy-to-deploy" with minimized infrastructure upgrades as mentioned in Section III-D. A variety of roads are introduced, including exclusive go-straight, exclusive turn, and multi-directional roads. Meanwhile, intersections come in different shapes and sizes, including one three-leg with four signal phases (the rightmost one in Fig.\ref{fig:Dublin}); the four-leg intersection is the majority, three have four phases, and the other has six phases; the most complex intersection is 5-leg with 6 phases (the third one from the left of Fig.\ref{fig:Dublin}). The scenario is extracted from the open data in \cite{gueriau2020quantifying} to simulate the real-world traffic in Dublin city. We extracted dynamic traffic generated from 10 AM for 400 seconds, consisting of 275 vehicles allowed to drive straight,  turn left or right at intersections. Each vehicle has a dedicated trip.

\subsection{Evaluation Metrics}
We evaluate the sustainable traffic improvements of each scenario using the following metrics:

\begin{itemize}
\item \textbf{Travel time (seconds)}:
Travel time of each vehicle is the time cost in the road network until finishing the designated trip. The average travel time is calculated on vehicles completing their trips in a scenario, which is the common measure to evaluate traffic efficiency \cite{wei2019survey}.
\item \textbf{Delay (seconds)}:
Delay is the difference between the actual travel time and the ideal travel time (i.e., time spent when driving at the maximum permitted speed) for each trip. This value indicates the space in which the traffic efficiency can be further optimized to its upper-bound. This metric could be more noticeable than travel time to reflect the improvement of traffic efficiency \cite{postigo2021effects}.
\item \textbf{Fuel consumption (l/100km)}:
Fuel consumption is the average amount of fuel consumed in liters every 100 kilometers traveled. The closer the vehicle speed is to the maximum speed limit we set, the more gentle change of acceleration, the less fuel consumption is likely to be achieved \cite{ahn2002estimating}. In our experiments, fuel consumption, as well as the CO\textsubscript{2} emission described later, is calculated using HBEFA3/PC\_G\_EU4 model (i.e., a gasoline-powered Euro norm 4-passenger car modeled using the HBEFA3\cite{keller2010handbook}), which is the default vehicle emission model in SUMO\footnote{https://sumo.dlr.de/docs/Models/Emissions.html}. This model mainly considers the instantaneous speed and acceleration of a vehicle.
\item \textbf{CO\textsubscript{2} emissions (g/km)}:
CO\textsubscript{2} emissions are measured by the average amount of carbon dioxide emitted in grams per kilometer traveled by all vehicles. As the primary component of greenhouse gas emissions, CO\textsubscript{2} emissions are required to be reduced to achieve sustainable traffic.
\item \textbf{Time-To-Collision (TTC)}: TTC is a widely-used safety indicator \cite{lee1976theory}, estimating the time required for a car to hit its preceding one. We use the default threshold of TTC in SUMO, 3 seconds\footnote{https://sumo.dlr.de/docs/Simulation/Output/SSM\_Device.html}, which means a possible collision is recognised when the time gap between the two adjacent cars is less than 3 seconds. The value of TTC is literally the total number of such possible rear-end collisions for a given time horizon.
\end{itemize}

\begin{table}[htbp]
\centering
\caption{Simulation settings of different vehicle types.}
\renewcommand\arraystretch{1.7}
\begin{threeparttable}
\begin{tabular}{|c|c|}
% \arrayrulecolor{blue}
\hline
\multicolumn{2}{|c|}{\textbf{All vehicles are CONNECTED}}\\
\hline
\textbf{HDV (Non-CAV)} & \textbf{CAV} \\ 
\hline
\multicolumn{1}{|l|}{\begin{tabular}{@{\labelitemi\hspace{\dimexpr\labelsep+0.5\tabcolsep}}l@{}}Can NOT be controlled by CoTV\\IDM car-following model \cite{treiber2017intelligent} \end{tabular}} & \multicolumn{1}{l|}{\begin{tabular}{@{\labelitemi\hspace{\dimexpr\labelsep+0.5\tabcolsep}}l@{}}Can be controlled by CoTV \tnote{1}\\IDM, if not controlled by CoTV\end{tabular}}  \\ 
\hline
\multicolumn{2}{|c|}{
$Penetration\, Rate = \frac{|CAV|}{|HDV| + |CAV|} \times 100\%$
} \\
\hline
\end{tabular}
\begin{tablenotes}
    \footnotesize
    \item[1] Denoted as CoTV* when all CAVs are controlled by our system. This scenario of 100\% penetration rate is a different case as CoTV only controls CAV that is the closest to the incoming intersection on each road.
\end{tablenotes}
\end{threeparttable}
\label{tab:vehicle}
\end{table}

\subsection{Compared Methods}
To evaluate the effectiveness of our system CoTV, the compared methods are described as follows:
\begin{itemize}
\item\textbf{Baseline}: 
This method is the baseline to evaluate the improvement of others. Traffic light signals have a static timing plan that does not change with the varying traffic, thus does not require V2X communications to collect vehicle information. All vehicles are HDV that are simulated by IDM car-following model \cite{treiber2017intelligent} as shown in Table \ref{tab:vehicle}, which is also used for simulating HDV in \cite{wu2017flow}. The Baseline scenario simulates most existing urban scenarios, which do not have any traffic light controllers and CAVs controlled by DRL. A cycle of the static traffic light signal plan contains four phases in order: Green-NS (green light for the flow N$\rightarrow$S and S$\rightarrow$N), Yellow-NS, Green-WE, and Yellow-WE. The duration of the green light is 40 seconds (default value in SUMO). The yellow light duration typically lasts from 3 to 6 seconds \cite{wei2019survey}, so we set 3 seconds as the yellow light duration, which is also the default setting in SUMO. Thus, the length of a cycle is 86 seconds (40+3+40+3). Besides, the Baseline of the Dublin scenario adopts the original traffic light signal plans. Their specific settings vary by different intersections. Green light phase duration ranges from 37 to 42 seconds, and yellow light phase lasts 3 seconds. Some of intersections have a short green light for turn-right with 6 seconds.
\item\textbf{FlowCAV}: 
FlowCAV \cite{wu2017flow} is a state-of-the-art DRL-based model to control the speed of a CAV to improve fuel efficiency and reduce emissions. Each CAV observes its preceding vehicle and then regulates its speed. The reward of a single CAV is evaluated globally by the average speed and acceleration of all vehicles. In this scenario, all traffic light signals are static. There is only one CAV agent per road, which leads the following vehicles on the same road.
\item\textbf{PressLight}: 
PressLight \cite{wei2019presslight} is a state-of-the-art DRL-based model to control traffic light signals to improve intersection throughput. The state of a traffic light controller includes the number of vehicles on the incoming roads and outgoing roads. The reward design utilizes the ``pressure" to improve intersection throughput, which is inspired by \cite{varaiya2013max}. All vehicles are HDV and connected, as shown in Table \ref{tab:vehicle}, which are periodically broadcast their up-to-date status (e.g., location, speed, acceleration), any agents within the communication range can aggregate them as the real-time traffic information.
\item\textbf{GLOSA}:
This is a optimization-based method for jointly controlling traffic light signals and CAVs. The GLOSA system\footnote{https://sumo.dlr.de/docs/Simulation/GLOSA.html} can adjust CAV speed considering the current traffic light phase and the current status of CAV. In our experiment, we combine it with adaptive traffic light controllers\footnote{https://sumo.dlr.de/docs/Simulation/Traffic\_Lights.html\#type\_actuated} to achieve joint control. Thus, phase switching is actuated after detecting a sufficient time gap between successive vehicles, resulting in various phase durations. It is worth noting that all vehicles in this scenario are CAVs.
\item{\textbf{I-CoTV}}:
I-CoTV combines independent policy training on the two types of agents as a common and straightforward way to develop MARL. There is no cooperation design between agents in either state or action, distinct from CoTV (action-independent MARL with cooperation schemes in the state exchange). Hence, the state of traffic light controllers involves two parts: its current signal phase and traffic on the roads it coordinates, not including any instantaneous vehicle information compared to CoTV. Correspondingly, the state of CAV agent only consists of the speed, acceleration, and location of itself and its preceding vehicle, without the current signal of the approaching traffic light from agent communication. Introducing I-CoTV aims to demonstrate that the efficient cooperation schemes of CoTV facilitate training convergence.
\item{\textbf{M-CoTV}}:
M-CoTV is the action-dependent MARL version of CoTV that trains the policies of traffic light controllers and CAVs considering both the action and state of another agent type within the range of one intersection. Introducing M-CoTV aims to demonstrate that CoTV takes advantage of the simplicity of action-independent MARL on policy training while efficiently achieving traffic improvements.
\item{\textbf{CoTV*}}: 
CoTV* remains all features of CoTV, except that in CoTV*, the traffic light controller interacts with \textit{all} CAVs instead of only the \textit{closest} one to the intersection on each incoming road. Introducing CoTV* aims to demonstrate the improvement of CoTV in alleviating scalability issues.
\end{itemize}

\section{Evaluation Results}

This section firstly discusses how CoTV performs in traffic efficiency and safety against four other compared methods under grid maps and Dublin scenario. Experiments with various CAV penetration rates are also conducted. Additionally, comparison of other MARL methods further demonstrates the effectiveness of CoTV on the cooperative control. Secondly, we show if CoTV can be efficiently deployed by comparing it with CoTV* in training time and traffic improvements. All numerical results shown are averaged from eighteen episodes.

\subsection{Traffic Efficiency \& Safety}

\subsubsection{Comparison with state-of-the-art methods}

Table \ref{tab:traf_perf} shows the traffic improvements of CoTV under 100\% CAV penetration rate, the same for FlowCAV and GLOSA (while 0\% CAV penetration rate for PressLight and Baseline scenario as no need for vehicle speed control).

\begin{table}[ht]
  \begin{center}
    \caption{Comparison of CoTV against Baseline and state-of-the-art methods. Percentage changes shown are compared to Baseline. The best achieved measurements are in bold.}
    \label{tab:traf_perf}
    \renewcommand{\arraystretch}{1.4}
    \begin{tabular}{m{1.2cm}|m{1.1cm}<{\centering}m{1.1cm}<{\centering}m{1.1cm}<{\centering}m{1.1cm}<{\centering}m{1.1cm}<{\centering}}
    \toprule
    \textbf{Method} & \textbf{Travel time (s)} & \textbf{Delay (s)} & \textbf{Fuel (l/100km)} & \textbf{CO\textsubscript{2} (g/km)} & \textbf{TTC}\\
    \hline
     & \multicolumn{5}{c}{\textbf{1x1 grid}} \\
    \hline
    \textbf{Baseline} & 59.67 & 18.76 & 9.29 & 216.05 & 354.00 \\
    \hline
    \multirow{2}{1.2cm}{\textbf{FlowCAV}} & 87.23 & 46.32 & 12.98 & 302.02 & 694.22 \\
    & +46.19\% & +146.91\% & +39.72\% & +39.79\% & +96.11\% \\
    \hline
    \multirow{2}{1.2cm}{\textbf{PressLight}} & 49.81 & 8.90 & 8.64 & 201.00 & 51.61  \\
    & -16.52\% & -52.56\% & -7.00\% & -6.97\% & -85.42\% \\
    \hline
    \multirow{2}{1.2cm}{\textbf{GLOSA}} & 50.95 & 10.03 & 8.65 & 201.34 & 65.83 \\
    & -14.61\% & -46.54\% & -6.89\% & -6.81\% & -81.40\% \\
    \hline
    \multirow{2}{1.2cm}{\textbf{CoTV}} & 48.42 & 7.50 & 8.44 & 196.42 & 30.11 \\
    & \textbf{-18.85\%} & \textbf{-60.02\%} & \textbf{-9.15\%} & \textbf{-9.09\%} & \textbf{-91.49\%} \\
    \hline
         & \multicolumn{5}{c}{\textbf{1x6 grid}} \\
    \hline
    \textbf{Baseline} & 89.99 & 33.13 & 9.54 & 221.97 & 1724.94 \\
    \hline
    \multirow{2}{1.2cm}{\textbf{FlowCAV}} & 172.12 & 118.30 & 17.60 & 409.44 & 5200.00 \\
    & +91.27\% & +257.08\% & +84.49\% & +84.46\% & +201.46\% \\
    \hline
    \multirow{2}{1.2cm}{\textbf{PressLight}} & 77.59 & 21.22 & 8.82 & 205.13 & 676.22 \\
    & -13.78\% & -35.95\% & -7.55\% & -7.59\% & -60.80\% \\
    \hline
    \multirow{2}{1.2cm}{\textbf{GLOSA}} & 68.91 & 12.05 & 7.59 & 176.49 & 252.94 \\
    & -23.42\% & -63.63\% & -20.44\% & -20.49\% & -85.34\% \\
    \hline
    \multirow{2}{1.2cm}{\textbf{CoTV}} & 65.56 & 8.70 & 7.27 & 169.19 & 68.28 \\
    & \textbf{-27.15\%} & \textbf{-73.74\%} & \textbf{-23.79\%} & \textbf{-23.78\%} & \textbf{-96.04\%} \\
    \hline
         & \multicolumn{5}{c}{\textbf{Dublin}} \\
    \hline
    \textbf{Baseline} & 59.33 & 29.17 & 10.98 & 255.53 & 1212.67 \\
    \hline
    \multirow{2}{1.2cm}{\textbf{FlowCAV}} & 59.39 & 29.43 & 11.16 & 259.70 & 1223.28 \\
    & +0.10\% & +0.89\% & +1.64\% & +1.63\% & +0.87\% \\
    \hline
    \multirow{2}{1.2cm}{\textbf{PressLight}} & 44.92 & 14.69 & 8.49 & 197.43 & 463.11 \\
    & -24.29\% & -49.64\% & -22.68\% & -22.74\% & -61.81\% \\
    \hline
    \multirow{2}{1.2cm}{\textbf{GLOSA}} & 45.40 & 15.06 & 8.46 & 196.92 & 545.50 \\
    & -23.48\% & -48.37\% & -22.95\% & -22.94\% & -55.02\% \\
    \hline
    \multirow{2}{1.2cm}{\textbf{CoTV}} & 41.76 & 11.48 & 7.97 & 185.42 & 195.94 \\
    & \textbf{-29.61\%} & \textbf{-60.64\%} & \textbf{-27.41\%} & \textbf{-27.44\%} & \textbf{-83.84\%} \\
    \bottomrule
    \end{tabular}
  \end{center}
\end{table}

\begin{figure}[htbp]
    \centering
    \includegraphics[width=0.95\linewidth]{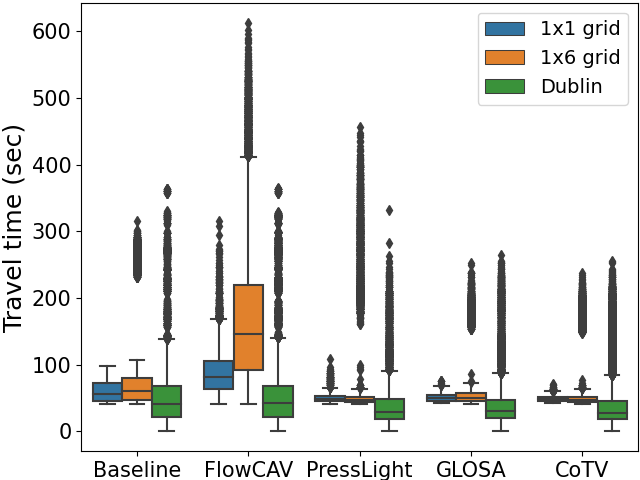}
    \caption{Travel time distributions for three test scenarios, comparing four methods. CoTV can reduce the travel time of all vehicles to be densely distributed with lower values than other methods.}
    \label{fig:traveltime}
\end{figure}

\begin{itemize}
    \item \textbf{Travel time \& delay}:
    As shown in Table \ref{tab:traf_perf}, CoTV achieves the shortest travel time with up to 30\% reduction compared to Baseline. PressLight and GLOSA achieve over 24\% and 23\% reduction, respectively. However, FlowCAV does not reduce travel time due to static traffic light plan and the absence of current traffic light signals, and the results in grid road maps are much worse than Baseline. The further improvement of CoTV demonstrates the advantages of cooperative traffic control compared with controlling traffic light signals only, meanwhile indicating that DRL-based approaches provide better adaptive traffic control than traditional approaches. Moreover, Fig.\ref{fig:traveltime} illustrates travel time of all vehicles can be reduced significantly and more densely distributed around a lower value under the three scenarios using CoTV, compared with other methods. The results in delay from Table \ref{tab:traf_perf} shows that CoTV reduces the travel time very close to its minimum possible value. Compared with other methods, CoTV can achieve up to about 74\% reduction in delay.
    \item \textbf{Environmental indicators}: CoTV brings the best results of fuel consumption and CO\textsubscript{2} emissions, both achieving over 27\% reduction shown in Table \ref{tab:traf_perf}. The reduced travel time of PressLight results in less fuel consumption. GLOSA obtains the second-best results due to the jointly optimised traffic light timings and vehicle speed. However, FlowCAV does not show any improvement on the two environmental indicators due to the complexity of urban scenarios containing intersections.
    \item \textbf{Traffic safety}: CoTV reduces TTC by over 96\%, as shown in Table \ref{tab:traf_perf}. PressLight and GLOSA improve traffic safety as well. However, there is a great difference in TTC between PressLight and CoTV under the 1x6 grid scenario, and the result of CoTV under Dublin scenario is much better than the other two methods. Conversely, FlowCAV hurts traffic safety under the grid maps but not in Dublin scenario. The more realistic urban scenario brings explicit complexity to enhance safety. This also highlights the advantages of CoTV using DRL-based methods for the cooperative control.
\end{itemize}

\begin{figure*}[htbp]
  \centering
  \subfloat[1x1 grid]{%
       \includegraphics[width=0.3\linewidth]{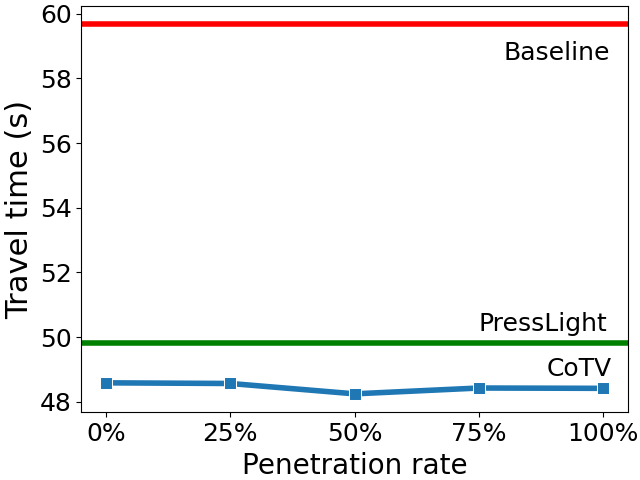}} \quad
  \subfloat[1x6 grid]{%
        \includegraphics[width=0.3\linewidth]{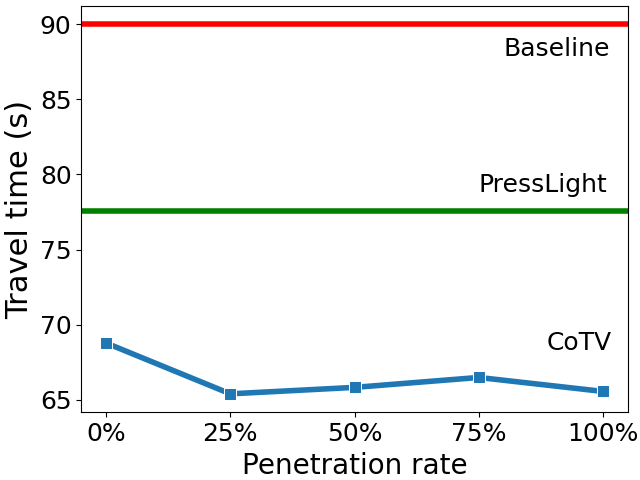}} \quad
  \subfloat[Dublin]{%
        \includegraphics[width=0.3\linewidth]{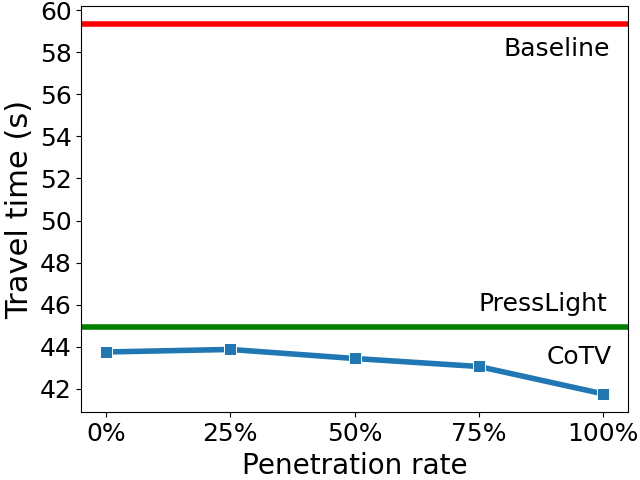}}
  \caption{The average travel time of CoTV under different penetration rates under both grid and Dublin scenarios. The average travel time obtained in Baseline and PressLight is also given for comparison. Travel time tends to decrease as the CAV penetration rate increases.}
  \label{fig:cav_deploy} 
\end{figure*}

\subsubsection{Robustness to varying CAV penetration rates}

Fig.\ref{fig:cav_deploy} shows that the travel time of CoTV tends to decrease as the CAV penetration rates increases under 1x1 and 1x6 grid maps and Dublin scenarios. Even under 0\% CAV penetration rate (i.e., the ratio of CAVs to all vehicles as shown in Table \ref{tab:vehicle}), the travel time that CoTV achieves is still less than Baseline and PressLight. Specifically, CoTV with 0\% CAV penetration rate implicates no vehicle speed control. In general, CoTV with CAV speed control can get better results, which demonstrates the effectiveness of cooperative traffic control. Similar results are shown in other metrics; thus, we do not present them to save space. This demonstrates the practicability of CoTV when deployed in a realistic mixed-autonomy scenario.

\subsubsection{Comparison with other MARL methods}

To further demonstrate the effectiveness of our CoTV system design on cooperative control, we compare CoTV with two other common MARL methods, I-CoTV (independent, without any cooperation schemes) and M-CoTV (action-dependent, with cooperation schemes in action and state). Results under Dublin scenario with full-autonomy traffic are shown in Table.\ref{tab:traf_perf_marl}. CoTV achieves the best results, while I-CoTV suffers from convergence issues, resulting in the worst traffic performance. M-CoTV fails to overcome high complexity from consideration of other agents' actions, which affects traffic improvements. In particular, the performance changes in fuel consumption and travel time are inconsistent in M-CoTV compared with I-CoTV. The training time of M-CoTV also increases by about 50\%.  In addition, referring to Table.\ref{tab:traf_perf}, M-CoTV and I-CoTV perform better than Baseline and FlowCAV but do not surpass PressLight and GLOSA.

\begin{table}[htbp]
  \begin{center}
    \caption{Comparison between I-CoTV (independent, without any cooperation schemes), M-CoTV (action-dependent, with cooperation schemes in action and state), and CoTV with full-autonomy traffic under Dublin scenario.}
    \label{tab:traf_perf_marl}
    \begin{tabular}{m{1.3cm}<{\centering}|m{1.1cm}<{\centering}|m{1.1cm}<{\centering}|m{1.1cm}<{\centering}|m{1.1cm}<{\centering}}
    \toprule
    \textbf{Method} & \textbf{Travel time (s)} & \textbf{Fuel (l/100km)} & \textbf{TTC}  & \textbf{Training time (h)}\\
    \midrule
    I-CoTV & 49.21 & 9.19 & 489.78 & 1.36\\
    \midrule
    M-CoTV & 47.53 & 10.44 & 660.08 & 2.00\\
    \midrule
    CoTV & 41.76 & 7.97 & 195.94 & 1.33\\
    \bottomrule
    \end{tabular}
  \end{center}
\end{table}

In summary, CoTV achieves the first three system goals, including reduced travel time, lower fuel consumption and CO\textsubscript{2} emissions, and longer time-to-collision. The cooperation schemes between CAV and traffic light controllers, which is the first contribution of this paper, can overcome the difficulties of DRL-based joint control in complex urban traffic scenarios.

\subsection{Scalability Improvement}

\begin{figure}[htbp] 
   \centering
    \includegraphics[width=0.95\linewidth]{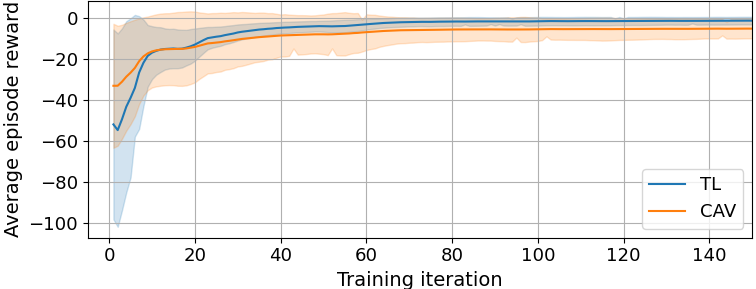}
  \caption{Evolution of the average episode reward for traffic light controllers (TL) and CAV agent of CoTV under Dublin scenario. The shade represents the standard deviation value. After DRL training on CoTV, the rewards for both types of agents can converge to higher values and smaller standard deviations than in the initial stage.}
  \label{fig:reward} 
\end{figure}

The second contribution of CoTV is the improvement of the multi-agent system scalability by reducing the number of CAV agents controlled. Compared with CoTV* that trains all possible CAVs, results from Table \ref{tab:traf_perf_scale} indicate that CoTV can reduce the training time by up to 44\%, while still having comparable (sometimes slightly better) improvement in both traffic efficiency and safety under Dublin scenario. Although CoTV* obtains better results under the two grid maps than CoTV, it is worth reminding that CoTV achieves this by only cooperating with the closest CAV on each incoming road for the traffic light controller. The closest CAV has the great potential to increase intersection throughput, which is similar to controlling the leading vehicle only for improving the traffic efficiency of a platoon \cite{lioris2016doubling}. The CAV as the leading vehicle is well controlled by CoTV, all its following vehicles are subsequently self-adjusted.
Moreover, Fig.\ref{fig:reward} indicates that two agent types of CoTV, traffic light controllers and CAVs, can be converged at a higher reward with a small standard deviation than the start after about 60 training iterations.
Thus, CoTV can alleviate scalability issues, while also not compromise traffic improvement. The last goal of system design, easier to deploy, is achieved.

\begin{table}[htbp]
  \begin{center}
    \caption{Comparison between CoTV and CoTV* (control all possible CAVs) under full-autonomy traffic.}
    % \hl{``fuel", ``travel time", ``TTC", and ``training time" these 4 columns only are good enough to draw our desired conclusion}
    \label{tab:traf_perf_scale}
    \arrayrulecolor{black}
    \begin{tabular}{m{0.4cm}<{\centering}m{1.0cm}<{\centering}|m{1.0cm}<{\centering}|m{1.4cm}<{\centering}|m{1.0cm}<{\centering}|m{1.0cm}<{\centering}}
    \toprule
    & \textbf{Method} & \textbf{Travel time (s)} & \textbf{Fuel (l/100km)} & \textbf{TTC}  & \textbf{Training time (h)}\\
    \midrule
    \multirow{2}{0.4cm}{\shortstack{\textbf{ 1x1} \\ \textbf{ grid }}} & CoTV* & 48.34 & 8.40 & 29.61 & 0.70\\
    & CoTV & 48.42 & 8.44 & 30.11 & 0.45\\
    \midrule
    \multirow{2}{0.4cm}{\shortstack{\textbf{ 1x6} \\ \textbf{ grid }}} & CoTV* & 65.34 & 7.11 & 61.78 & 2.57\\
    & CoTV & 65.56 & 7.27 & 68.28 & 1.65\\
    \midrule
    \multirow{2}{0.4cm}{\textbf{Dublin}} & CoTV* & 43.42 & 7.98 & 219.67 & 2.37\\
    & CoTV & 41.76 & 7.97 & 195.94 & 1.33\\
    \bottomrule
    \end{tabular}
  \end{center}
\end{table}

\subsection{Discussion}

\begin{figure}[ht!]
    \centering
    \includegraphics[width=0.7\linewidth]{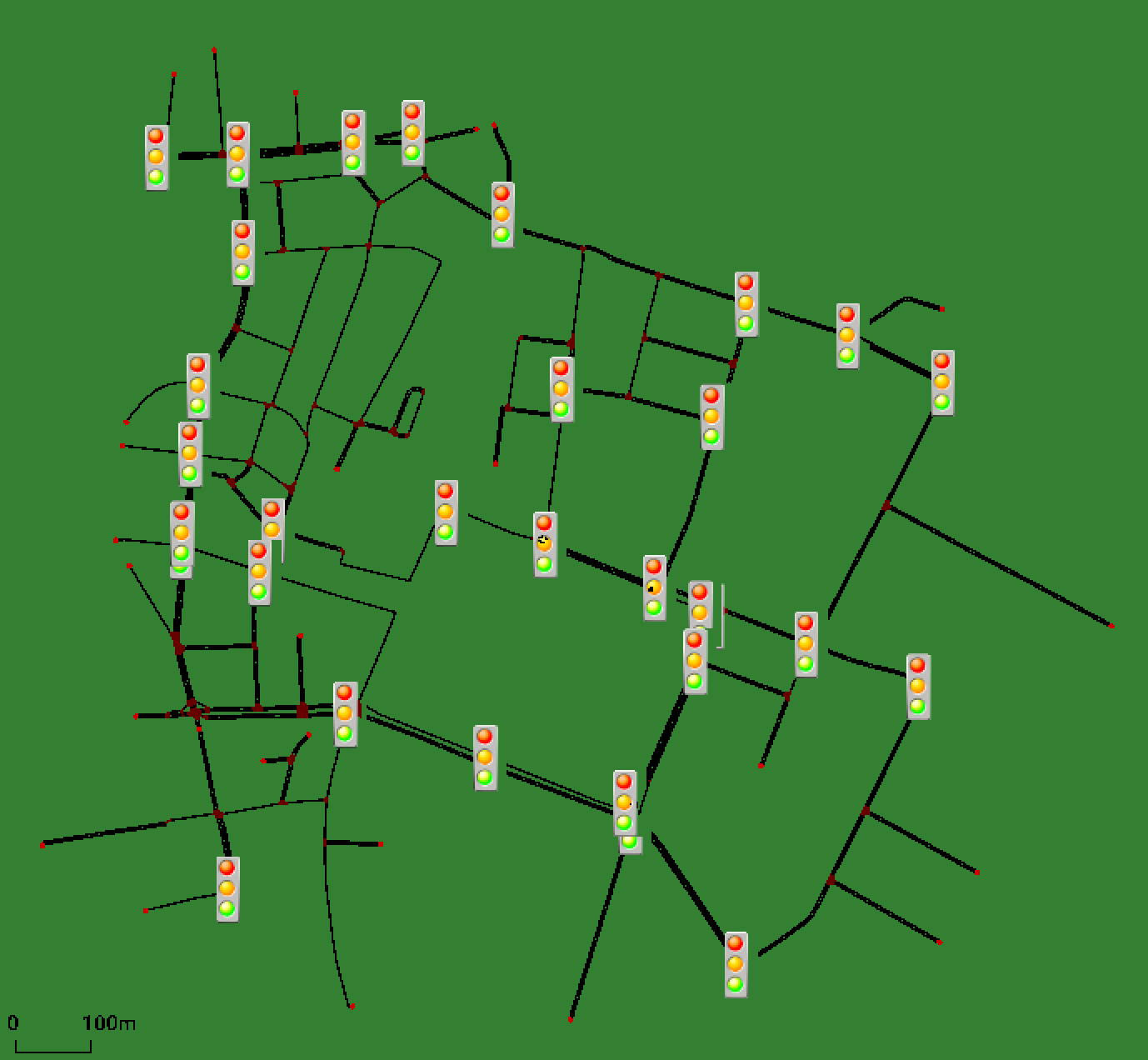}
    \caption{The selected dense urban scenario in the city centre of Dublin. There are 119 intersections in total, including 31 signalized intersections. 321 vehicles are generated from 10 AM in 400 seconds.}
    \label{fig:Dublin_1kmsignal}
\end{figure}

To further explore the deployment options of CoTV, we conduct experiments under a relatively large and dense urban scenario in Dublin city centre, which traditionally requires sophisticated coordination between adjacent traffic light controllers. The selected area covers nearly 1 $km^2$ with 31 signalized intersections, as shown in Fig.\ref{fig:Dublin_1kmsignal}. These intersections with different road shapes and traffic light signal cycles/phases are all controlled by CoTV. Table.\ref{tab:perf_1km} shows the traffic performance under this dense Dublin scenario under 100\% CAV penetration rate. Although CoTV can get converged and obtain the best results in all evaluation metrics, which shows that CoTV can be deployed in both major and minor junctions, we still need further studies to find the optimal selection of key intersections to control to avoid costly deployment on all urban junctions.

\begin{table}[ht]
  \begin{center}
    \caption{Traffic performance under a dense Dublin scenario. Percentage changes shown are compared to Baseline. The best achieved measurements are in bold.}
    \label{tab:perf_1km}
    \renewcommand{\arraystretch}{1.4}
    \begin{tabular}{m{1.2cm}|m{1.1cm}<{\centering}m{1.1cm}<{\centering}m{1.1cm}<{\centering}m{1.1cm}<{\centering}m{1.1cm}<{\centering}}
    \toprule
    \textbf{Method} & \textbf{Travel time (s)} & \textbf{Delay (s)} & \textbf{Fuel (l/100km)} & \textbf{CO\textsubscript{2} (g/km)} & \textbf{TTC}\\
    \hline
    \textbf{Baseline} & 125.40 & 78.84 & 11.22 & 261.09 & 2344.50 \\
    \hline
    \multirow{2}{1.2cm}{\textbf{FlowCAV}} & 127.71 & 81.14 & 11.06 & 257.03 & 1947.44 \\
    & +1.84\% & +2.92\% & -1.43\% & -1.45\% & -16.94\% \\
    \hline
    \multirow{2}{1.2cm}{\textbf{PressLight}} & 114.60 & 68.15 & 9.70 & 225.59 & 1800.83 \\
    & -8.61\% & -13.56\% & -13.55\% & -13.60\% & -23.19\% \\
    \hline
    \multirow{2}{1.2cm}{\textbf{GLOSA}} & 104.21 & 57.64 & 8.50 & 197.70 & 1193.61 \\
    & -16.90\% & -26.89\% & -24.24\% & -24.28\% & -49.09\% \\
    \hline
    \multirow{2}{1.2cm}{\textbf{CoTV}} & 103.18 & 56.60 & 8.40 & 195.39 & 787.29 \\
    & \textbf{-17.72\%} & \textbf{-28.21\%} & \textbf{-25.13\%} & \textbf{-25.16\%} & \textbf{-66.42\%} \\
    \bottomrule
    \end{tabular}
  \end{center}
\end{table}

\section{Conclusions and Future Work}

This paper proposes a multi-agent DRL system, CoTV, to control traffic light signals and CAV cooperatively for achieving sustainable traffic goals. CoTV can significantly improve traffic efficiency (i.e., travel time, fuel consumption, and CO\textsubscript{2} emissions) as well as traffic safety (i.e., time-to-collision), which outperforms other DRL-based systems that control either traffic light signal or vehicle speed, and traditional joint control method. Moreover, the traffic light controllers in our CoTV utilize V2I communications infrastructure to only cooperate with the closest CAV (i.e., as the leader of a platoon) on each incoming road for alleviating scalability issue of multi-agent systems. This also eases the deployment and achieves the training process to converge within a moderate number of iterations. Experiments in various grid maps and realistic urban scenarios demonstrate the effectiveness of CoTV. Compared to the Baseline, CoTV can save up to 28\% in fuel consumption and CO\textsubscript{2} while reducing travel time by up to 30\%. The robustness of CoTV is also validated under different penetration rates of CAV.

As future works, we plan to improve the robustness of our CoTV system to more practical scenarios. Firstly, we will relax the assumption that all vehicles are connected using V2X communications. Secondly, we will improve CoTV to be resilient to varying vehicular network conditions (e.g., latency, packet loss, bandwidth, etc.). Our long-term goal is to tackle the scalability issues of applying cooperative MARL algorithms (e.g., COMA) in complex urban traffic scenarios.

% In the future, we will further investigate the potential problems when assessing CoTV under various realistic scenarios, including 1) relaxing the assumption that all vehicles are connected vehicles; 2) improving the robustness of the system to eliminate the effects of packet loss and delays in real vehicular networks; 3) optimizing the algorithm used and analyzing other multi-agent DRL algorithms as alternatives.

\bibliographystyle{IEEEtran}
\bibliography{references.bib}

\begin{IEEEbiography}[{\includegraphics[width=1in,height=1.25in,clip,keepaspectratio]{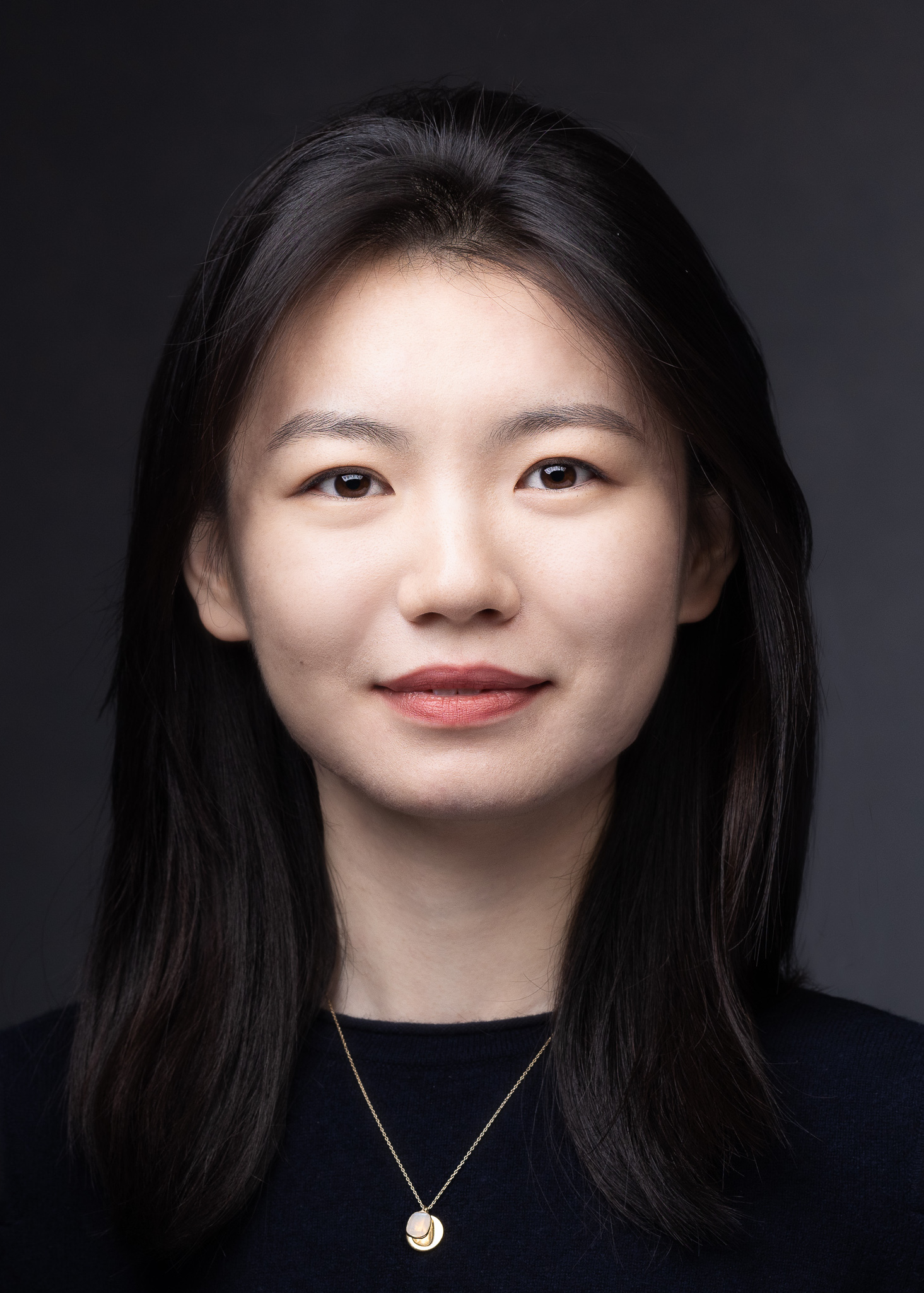}}]{Jiaying Guo} (Student Member, IEEE)
is a Ph.D. student at the School of Computer Science, University College Dublin, Ireland. She received the B.Sc. degree from University College Dublin and the B.Eng. degree from Beijing University of Technology, China, in 2020. Her research interests include multi-agent systems of intelligent transportation, reinforcement learning, mix-autonomy traffic, and connected autonomous vehicles.
\end{IEEEbiography}

\begin{IEEEbiography}[{\includegraphics[width=1in,height=1.25in,clip,keepaspectratio]{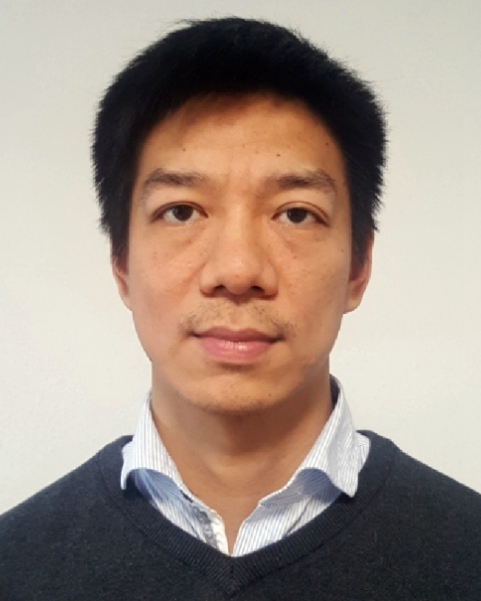}}]{Long Cheng} is a Full Professor in the School of Control and Computer Engineering at North China Electric Power University in Beijing, and also a Visiting Professor at Insight SFI Research Centre for Data Analytics in Dublin.  He received the B.E. from Harbin Institute of Technology, China in 2007, M.Sc from University of Duisburg-Essen, Germany in 2010 and Ph.D from National University of Ireland Maynooth in 2014. He was an Assistant Professor at Dublin City University, and a Marie Curie Fellow at University College Dublin. He also has worked at organizations such as Huawei Technologies Germany, IBM Research Dublin, TU Dresden and TU Eindhoven. He has published around 60 papers in journals and conferences like TPDS, TON, TC, TSC, TASE, TCAD, TCC, TBD, TITS, TVLSI, JPDC, IEEE Network, CIKM, ICPP, CCGrid and Euro-Par etc. His research focuses on distributed systems,  deep learning, cloud computing and process mining. Prof Cheng is a Senior Member of the IEEE and an associate editor of the Journal of Cloud Computing.
\end{IEEEbiography}

\begin{IEEEbiography}[{\includegraphics[width=1in,height=1.25in,clip,keepaspectratio]{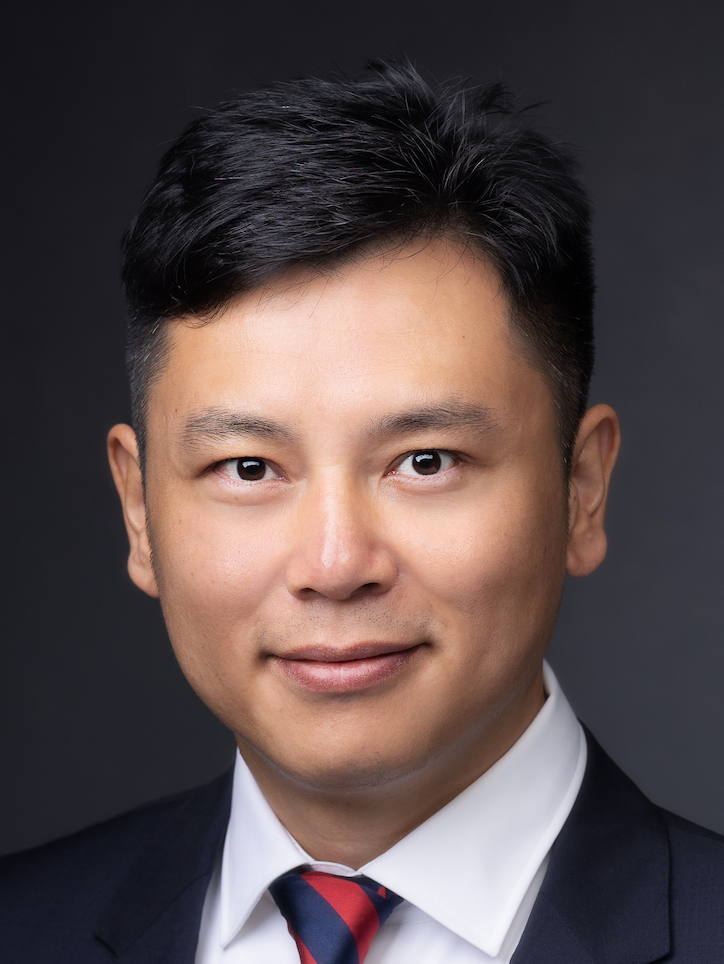}}]{Shen Wang} (Member, IEEE)
is currently an Assistant Professor with the School of Computer Science, University College Dublin, Ireland. He received the M.Eng. degree from Wuhan University, China, and the Ph.D. degree from Dublin City University, Ireland. Dr. Wang is a member of the IEEE and has been involved with several EU projects as a co-PI, WP and Task leader in big trajectory data streaming for air traffic control and trustworthy AI for intelligent cybersecurity systems. Some key industry partners of his applied research are IBM Research Brazil, Boeing Research and Technology Europe, and Huawei Ireland Research Centre. He is the recipient of the IEEE Intelligent Transportation Systems Society Young Professionals Travelling Fellowship 2022. His research interests include connected autonomous vehicles, explainable artificial intelligence, and security and privacy for mobile networks.
\end{IEEEbiography}

\vfill

\end{document}